%% file: main.tex
\documentclass[10pt,twocolumn,letterpaper]{article}
\pdfoutput=1
\usepackage[pagenumbers]{cvpr} 

\usepackage{amssymb}
\usepackage{booktabs}
\usepackage{multirow}

\input{preamble}

\definecolor{cvprblue}{rgb}{0.21,0.49,0.74}
\usepackage{hyperref}
\hypersetup{pagebackref,breaklinks,colorlinks,citecolor=cvprblue}

\def\paperID{*****} 
\def\confName{CVPR}
\def\confYear{2024}

\title{\vspace{-3mm}AnomalyGPT: Detecting Industrial Anomalies Using \\ Large Vision-Language Models\vspace{-3mm}}

\author{
Zhaopeng Gu$^{1,2}$~~~~
Bingke Zhu$^{1,3,4}$~~~~
Guibo Zhu$^{1,2,4}$~~~~\\
Yingying Chen$^{1,3,4}$~~~~
Ming Tang$^{1,2}$~~~~
Jinqiao Wang$^{1,2,3,4}$\\
  $^{1}$~Foundation Model Research Center, Institute of Automation, \\ Chinese Academy of Sciences, Beijing, China \\ 
  $^{2}$~University of Chinese Academy of Sciences, Beijing, China\\
  $^{3}$~Objecteye Inc., Beijing, China\\
  $^{4}$~Wuhan AI Research, Wuhan, China\\
  {\tt\small  guzhaopeng2023@ia.ac.cn} \\
  {\tt\small \{bingke.zhu,gbzhu,yingying.chen,tangm,jqwang\}@nlpr.ia.ac.cn}\vspace{-3mm}
}

\begin{document}
\maketitle

\begin{abstract}\vspace{-3mm}
  Large Vision-Language Models~(LVLMs) such as MiniGPT-4 and LLaVA have demonstrated the capability of understanding images and achieved remarkable performance in various visual tasks. Despite their strong abilities in recognizing common objects due to extensive training datasets, they lack specific domain knowledge and have a weaker understanding of localized details within objects, which hinders their effectiveness in the Industrial Anomaly Detection~(IAD) task. On the other hand, most existing IAD methods only provide anomaly scores and necessitate the manual setting of thresholds to distinguish between normal and abnormal samples, which restricts their practical implementation. In this paper, we explore the utilization of LVLM to address the IAD problem and propose AnomalyGPT, a novel IAD approach based on LVLM. We generate training data by simulating anomalous images and producing corresponding textual descriptions for each image. We also employ an image decoder to provide fine-grained semantic and design a prompt learner to fine-tune the LVLM using prompt embeddings. Our AnomalyGPT eliminates the need for manual threshold adjustments, thus directly assesses the presence and locations of anomalies. Additionally, AnomalyGPT supports multi-turn dialogues and exhibits impressive few-shot in-context learning capabilities. With only one normal shot, AnomalyGPT achieves the state-of-the-art performance with an accuracy of 86.1\%, an image-level AUC of 94.1\%, and a pixel-level AUC of 95.3\% on the MVTec-AD dataset. Code is available at \url{https://github.com/CASIA-IVA-Lab/AnomalyGPT}.
    \vspace{-3mm}
\end{abstract}

\begin{figure}[t]
    \centering
    \includegraphics[width=0.95\columnwidth]{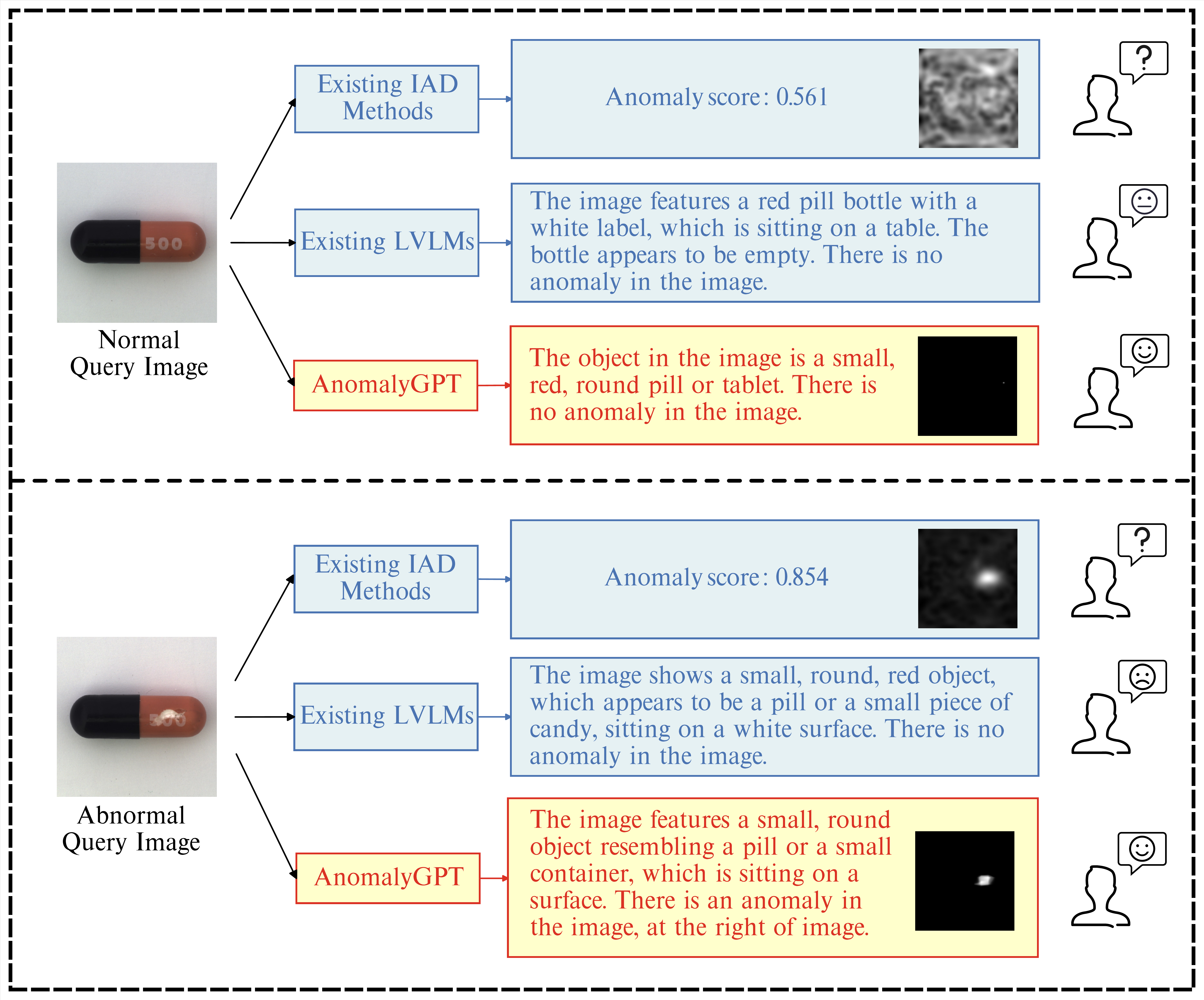} 
    \caption{Comparison between our AnomalyGPT, existing IAD methods and existing LVLMs. Existing IAD methods can only provide anomaly scores and need manually threshold setting, while existing LVLMs cannot detect anomalies in the image. AnomalyGPT can not only provide information about the image but also indicate the presence and location of anomaly.}
    \label{fig:IAD}
\end{figure}

\begin{table*}[]
    \centering
	\small
    \begin{tabular}{@{}cccccc@{}}
        \toprule
        Methods                & Few-shot learning & Anomaly score & Anomaly localization & Anomaly judgement & Multi-turn dialogue \\ \midrule
        Traditional IAD methods &                   & $\checkmark$  & $\checkmark$ &                    &                     \\
        Few-shot IAD methods    & $\checkmark$      & $\checkmark$  & $\checkmark$ &                    &                     \\
        LVLMs                  & $\checkmark$      &               &              &                    & $\checkmark$        \\
        \textbf{AnomalyGPT~(ours)}             & $\checkmark$      & $\checkmark$  & $\checkmark$ & $\checkmark$       & $\checkmark$        \\ \bottomrule
    \end{tabular}
    \caption{
        Comparison between our AnomalyGPT and existing methods across various functionalities. The ``Traditional~IAD~methods'' in the table refers to ``one-class-one-model'' methods such as PatchCore~\cite{roth2022towards}, InTra~\cite{pirnay2022inpainting}, and PyramidFlow~\cite{lei2023pyramidflow}.   ``Few-shot IAD methods'' refers to methods that can perform few-shot learning like RegAD~\cite{huang2022registration}, Graphcore~\cite{xie2023pushing}, and WinCLIP~\cite{wang2023visionllm}. ``LVLMs'' represents general large vision-language models like MiniGPT-4~\cite{zhu2023minigpt}, LLaVA~\cite{liu2023visual}, and PandaGPT~\cite{su2023pandagpt}. ``Anomaly score'' in the table represents just providing scores for anomaly detection, while ``Anomaly judgement'' indicates directly assessing the presence of anomaly.
          }
    \label{table1}
\end{table*}

\section{Introduction}
Large Language Models~(LLMs) like GPT-3.5~\cite{ouyang2022training} and LLaMA~\cite{touvron2023llama} have demonstrated remarkable performance on a range of Natural Language Processing~(NLP) tasks. More recently, novel methods including MiniGPT-4~\cite{zhu2023minigpt}, BLIP-2~\cite{li2023blip}, and PandaGPT~\cite{su2023pandagpt} have further extended the ability of LLMs into visual processing by aligning visual features with text features, bringing a significant revolution in the domain of Artificial General Intelligence~(AGI). While LVLMs are pre-trained on amounts of data sourced from the Internet, their domain-specific knowledge is relatively limited and they lack sensitivity to local details within objects, which restricts their potentiality in IAD task.

IAD task aims to detect and localize anomalies in industrial product images. Due to the rarity and unpredictability of real-world samples, models are required to be  trained only on normal samples and distinguish anomalous samples that deviate from normal samples. Current IAD methods~\cite{jeong2023winclip, huang2022registration, you2022unified} typically only provide anomaly scores for test samples and require manually specification of thresholds to distinguish between normal and anomalous instances for each class of items, which is not suitable for real production environments.

As illustrated in Figure~\ref{fig:IAD} and Table~\ref{table1}, neither existing IAD methods nor LVLMs can address IAD problem well, so we introduce AnomalyGPT, a novel IAD approach based on LVLM. AnomalyGPT can detect the presence and location of anomalies without the need for manual threshold settings. Moreover, our method can provide information about the image and allows for interactive engagement, enabling users to ask follow-up questions based on their needs and the provided answers. AnomalyGPT can also perform in-context learning with a small number of normal samples, enabling swift adaptation to previously unseen objects.

Specifically, we focus on fine-tuning the LVLM using synthesized anomalous visual-textual data, integrating IAD knowledge into the model. However, direct training with IAD data presents numerous challenges. The first is data scarcity. Methods like LLaVA~\cite{liu2023visual} and PandaGPT~\cite{su2023pandagpt} are pre-trained on 160k images with corresponding multi-turn dialogues. However, existing IAD datasets~\cite{bergmann2019mvtec,zou2022spot} contain only a few thousand samples, rendering direct fine-tuning easy to overfitting and catastrophic forgetting. To address this, we use prompt embeddings to fine-tune the LVLM instead of parameter fine-tuning. Additional prompt embeddings are added after image inputs, introducing supplementary IAD knowledge into the LVLM. The second challenge relates to fine-grained semantic. We propose a lightweight, visual-textual feature-matching-based decoder to generate pixel-level anomaly localization results. The decoder's outputs are introduced to the LVLM along with the original test images through prompt embeddings, which allows the LVLM to utilize both the raw image and the decoder's outputs to make anomaly determinations, improving the accuracy of its judgments.

Experimentally, we conduct extensive experiments on the MVTec-AD~\cite{bergmann2019mvtec} and VisA~\cite{zou2022spot} datasets. With unsupervised training on the MVTec-AD dataset, we achieve an accuracy of 93.3\%, an image-level AUC of 97.4\%, and a pixel-level AUC of 93.1\%. When one-shot transferred to the VisA dataset, we achieve an accuracy of 77.4\%, an image-level AUC of 87.4\%, and a pixel-level AUC of 96.2\%. Conversely, after unsupervised training on the VisA dataset, one-shot transferred to the MVTec-AD dataset result in an accuracy of 86.1\%, an image-level AUC of 94.1\%, and a pixel-level AUC of 95.3\%. 

Our contributions are summarized as follows:

\begin{itemize}
    \item We present the pioneering utilization of LVLM for addressing IAD task. Our method not only detects and locates anomaly without manually threshold adjustments but also supports multi-round dialogues. To the best of our knowledge, we are the first to successfully apply LVLM to the domain of industrial anomaly detection.
    \item The lightweight, visual-textual feature-matching-based decoder in our work addresses the limitation of the LLM's weaker discernment of fine-grained semantic and alleviates the constraint of LLM's restricted ability to solely generate text outputs.
    \item We employ prompt embeddings for fine-tuning and train our model concurrently with the data utilized during LVLM pre-training, thus preserving the LVLM's inherent capabilities and enabling multi-turn dialogues.
    \item Our method retains robust transferability and is capable of engaging in in-context few-shot learning on new datasets, yielding outstanding performance.
\end{itemize}

\section{Related Work}

\noindent\textbf{Industrial Anomaly Detection:}
Existing IAD methods can be categorized into reconstruction-based and feature embedding-based approaches. Reconstruction-based methods primarily aim to reconstruct anomalous samples to their corresponding normal counterparts and detect anomalies by calculating the reconstruction error. RIAD~\cite{zavrtanik2021reconstruction}, SCADN~\cite{yan2021learning}, InTra~\cite{pirnay2022inpainting} and AnoDDPM~\cite{wyatt2022anoddpm} employ different reconstruction network architectures, ranging from autoencoder and Generative Adversarial Network~(GAN) to Transformer and diffusion model.

Feature embedding-based methods focus on modeling the feature embeddings of normal samples. Approaches such as PatchSVDD~\cite{yi2020patch} aim to find a hypersphere that tightly encapsulates normal samples. Cflow-AD~\cite{gudovskiy2022cflow} and PyramidFlow~\cite{lei2023pyramidflow} use normalizing flows to project normal samples onto a Gaussian distribution. PatchCore~\cite{roth2022towards} and CFA~\cite{lee2022cfa} establish a memory bank of patch embeddings from normal samples and detect anomalies by measuring the distance between a test sample embedding and its nearest normal embedding in the memory bank.

These methods typically follow the ``one-class-one-model'' learning paradigm, requiring plentiful normal samples for each object class to learn its distribution, making them impractical for novel object categories and less suitable for dynamic production environments. In contrast, our method facilitates in-context learning for novel object categories, enabling inference with only few normal samples.

\begin{figure*}[t]
    \centering
    \includegraphics[width=0.95\textwidth]{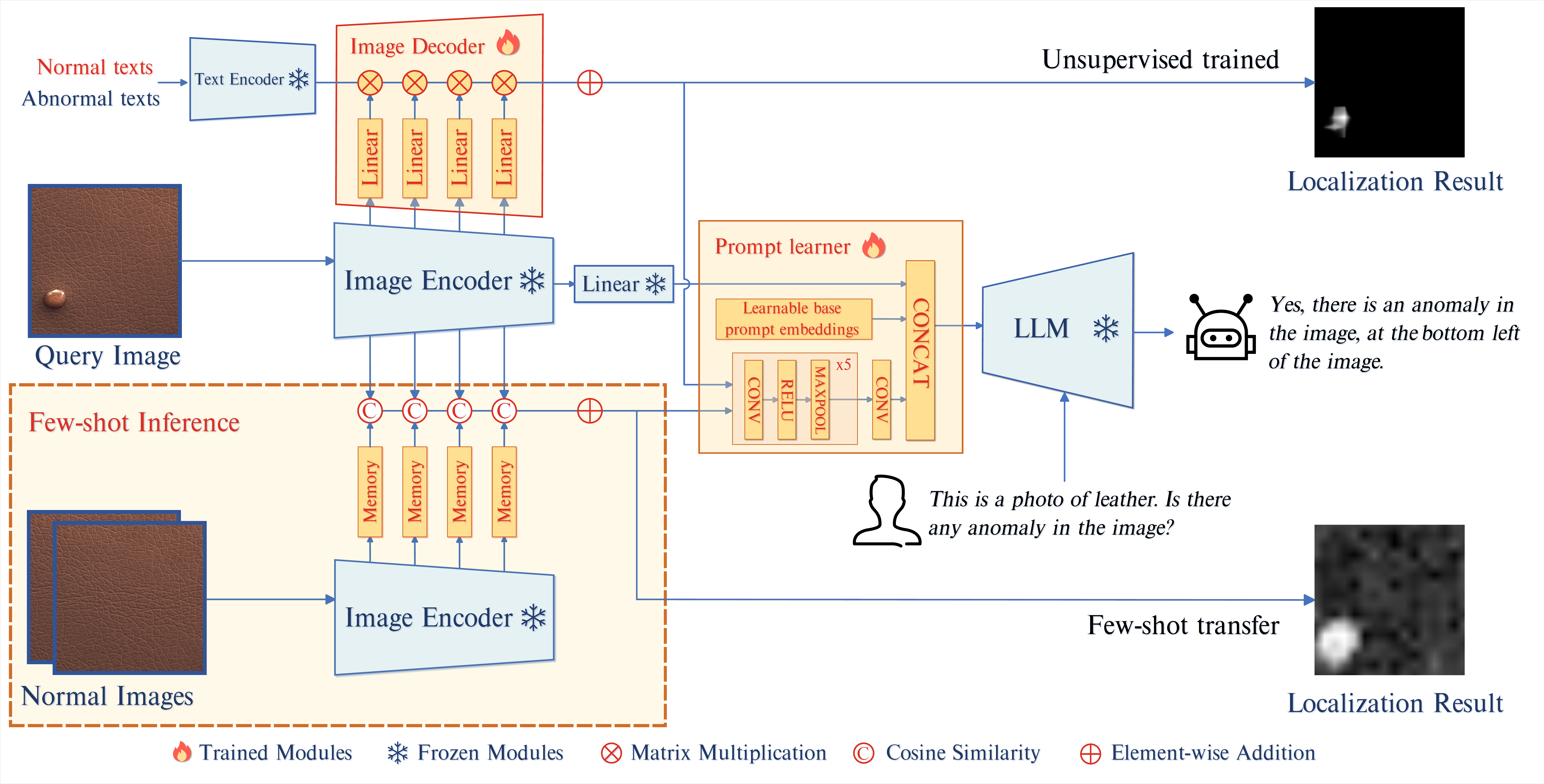} 
    \caption{The architecture of AnomalyGPT. The query image is passed to the frozen image encoder and the patch-level features extracted from intermediate layers are fed into image decoder to compute their similarity with normal and abnormal texts to obtain localization result. The final features extracted by the image encoder are fed to a linear layer and then passed to the prompt learner along with the localization result. The prompt learner converts them into prompt embeddings suitable for input into the LLM together with user text inputs. In few-shot setting, the patch-level features from normal samples are stored in memory banks and the localization result can be obtained by calculating the distance between query patches and their most similar counterparts in the memory bank.}
    \label{fig:structure}
\end{figure*}

\noindent\textbf{Zero-/Few-shot Industrial Anomaly Detection:}
Recent efforts have focused on methods utilizing minimal normal samples to accomplish IAD task. PatchCore~\cite{roth2022towards} constructs a memory bank using only a few normal samples, resulting in a noticeable performance decline. RegAD~\cite{huang2022registration} trained an image registration network to align test images with normal samples, followed by similarity computation for corresponding patches. WinCLIP~\cite{jeong2023winclip} leveraged CLIP~\cite{radford2021learning} to compute similarity between images and textual descriptions representing normal and anomalous semantics, distinguishing anomalies based on their relative scores.
\indent However, these methods can only provide anomaly scores for test samples during inference. To distinguish normal samples from anomalous ones, it's necessary to experimentally determine the optimal threshold on a test set, which contradicts the original intent of IAD task that only utilize normal data. For instance, while PatchCore~\cite{roth2022towards} achieves an image-level AUC of 99.3\% on MVTec-AD in unsupervised setting, its accuracy drops to 79.76\% when using a unified threshold for inference. The detailed experimental results and analyses can be found in Appendix A. Our method, in contrast, enables the LVLM to directly assess test samples for the presence of anomalies and pinpoint their locations, demonstrating enhanced practicality.

\noindent\textbf{Large Vision-Language Models:}
LLMs, traditionally successful in NLP, are now explored for visual tasks. BLIP-2~\cite{li2023blip} leverages Q-Former to input visual features from Vision Transformer~\cite{dosovitskiy2020image} into the Flan-T5~\cite{chung2022scaling} model. MiniGPT-4~\cite{zhu2023minigpt} connects the image segment of BLIP-2 and the Vicuna~\cite{chiang2023vicuna} model with a linear layer, performing a two-stage fine-tuning process using extensive image-text data. PandaGPT~\cite{su2023pandagpt} establishes a connection between ImageBind~\cite{girdhar2023imagebind} and the Vicuna~\cite{chiang2023vicuna} model via a linear layer, allowing for multi-modal input. These approaches showcase the potential of LLM-based polymathic models.

However, as mentioned earlier, these models are trained on general data and lack domain-specific expertise. In this paper, through the utilization of simulated anomaly data, image decoder and prompt embeddings, AnomalyGPT is introduced as an novel approach that achieves IAD task without the need for manually specified thresholds, while also enabling few-shot in-context learning. Table~\ref{table1} illustrates a comparison between AnomalyGPT and existing methods across various functionalities.

\section{Method}
AnomalyGPT is a novel conversational IAD vision-language model, primarily designed for detecting anomalies in images of industrial artifacts and pinpointing their positions. We leverage a pre-trained image encoder and a LLM to align IAD images and their corresponding textual descriptions via simulated anomaly data. We introduce a decoder module and a prompt learner module to enhance IAD performance and achieve pixel-level localization output. Employing prompt tuning and alternate training with pre-training data preserves the LLM's transferability and prevents catastrophic forgetting. Our method exhibits robust few-shot transfer capability, enabling anomaly detection and localization for previously unseen items with merely one normal sample provided.

\subsection{Model Architecture}
Figure \ref{fig:structure} illustrates the comprehensive architecture of AnomalyGPT. Given a query image $x \in \mathbb{R}^{H\times W\times C}$, the final features $F_{img} \in \mathbb{R}^{C_1}$ extracted by the image encoder are passed through the linear layer to obtain the image embedding $E_{img} \in \mathbb{R}^{C_{emb}}$, which is then fed into the LLM. In unsupervised setting, the patch-level features extracted by intermediate layers of image encoder are fed into the decoder together with text features to generate pixel-level anomaly localization results. In few-shot setting, the patch-level features from normal samples are stored in memory banks and the localization result can be obtained by calculating the distance between query patches and their most similar counterparts in the memory bank. The localization results is subsequently transformed into prompt embeddings through the prompt learner, serving as a part of LLM input. The LLM leverages image input, prompt embeddings, and user-provided textual input to detect anomalies and identify their locations, thus generating responses for the user.

\subsection{Decoder and Prompt Learner}
\noindent\textbf{Decoder}
To achieve pixel-level anomaly localization, we employ a lightweight feature-matching-based image decoder that supports both unsupervised IAD and few-shot IAD. The design of the decoder is primarily inspired by PatchCore~\cite{roth2022towards}, WinCLIP~\cite{jeong2023winclip}, and APRIL-GAN~\cite{chen2023zero}.

As illustrated in the upper part of Figure \ref{fig:structure}, we partition the image encoder into 4 stages and obtain the intermediate patch-level features extracted by every stage $F^{i}_{patch} \in \mathbb{R}^{H_i\times W_i\times C_i}$, where $i$ indicates the $i$-th stage. Following the idea from WinCLIP~\cite{jeong2023winclip}, a natural approach is to compute the similarity between $F^{i}_{patch}$~and the text features $F_{text} \in \mathbb{R}^{2\times C_{text}}$ respectively representing normality and abnormality. Detailed texts representing normal and abnormal cases are presented in Appendix B. However, since these intermediate features have not undergone the final image-text alignment, they cannot be directly compared with text features. To address this, we introduce additional linear layers to project these intermediate features to $\tilde{F}^{i}_{patch} \in \mathbb{R}^{H_i\times W_i\times C_{text}}$, and align them with text features representing normal and abnormal semantics. The localization result $M \in \mathbb{R}^{H\times W}$ can be obtained by Eq.~(\ref{eq:localization result}):

\begin{equation}
    M = Upsample\left(\sum_{i=1}^{4}{softmax(\tilde{F}^{i}_{patch}F_{text}^{T})}\right). \label{eq:localization result}
\end{equation}

For few-shot IAD, as illustrated in the lower part of Figure \ref{fig:structure}, we utilize the same image encoder to extract intermediate patch-level features from normal samples and store them in memory banks $B^i \in \mathbb{R}^{N\times C_i}$, where $i$ indicates the $i$-th stage. For patch-level features $F^{i}_{patch} \in \mathbb{R}^{H_i\times W_i\times C_i}$, we calculate the distance between each patch and its most similar counterpart in the memory bank, and the localization result $M \in \mathbb{R}^{H\times W}$ can be obtained by Eq.~(\ref{eq:few-shot localization result}):

\begin{equation}
    M = Upsample\left(\sum_{i=1}^{4}{\left(1 - max(F^{i}_{patch} \cdot {B^i}^T)\right)}\right). \label{eq:few-shot localization result}
\end{equation}

\noindent\textbf{Prompt Learner}
To leverage fine-grained semantic from images and maintain semantic consistency between LLM and decoder outputs, we introduce a prompt learner that transforms the localization result into prompt embeddings. Additionally, learnable base prompt embeddings, unrelated to decoder outputs, are incorporated into the prompt learner to provide extra information for the IAD task. Finally, these embeddings, along with the original image information, are fed into the LLM.

As illustrated in Figure~\ref{fig:structure}, the prompt learner consists of the learnable base prompt embeddings $E_{base} \in \mathbb{R}^{n_1 \times C_{emb}} $ and a convolutional neural network. The network converts the localization result $M \in R^{H\times W} $ into $n_2$ prompt embeddings $E_{dec} \in \mathbb{R}^{n_2 \times C_{emb}}$. $E_{base}$ and $E_{dec}$ form a set of $n_1+n_2$ prompt embeddings $E_{prompt} \in \mathbb{R}^{(n_1+n_2) \times C_{emb}}$ that are combined with the image embedding into the LLM.

\subsection{Data for Image-Text Alignment}
\begin{figure}[h]
    \centering
    \includegraphics[width=0.9\columnwidth]{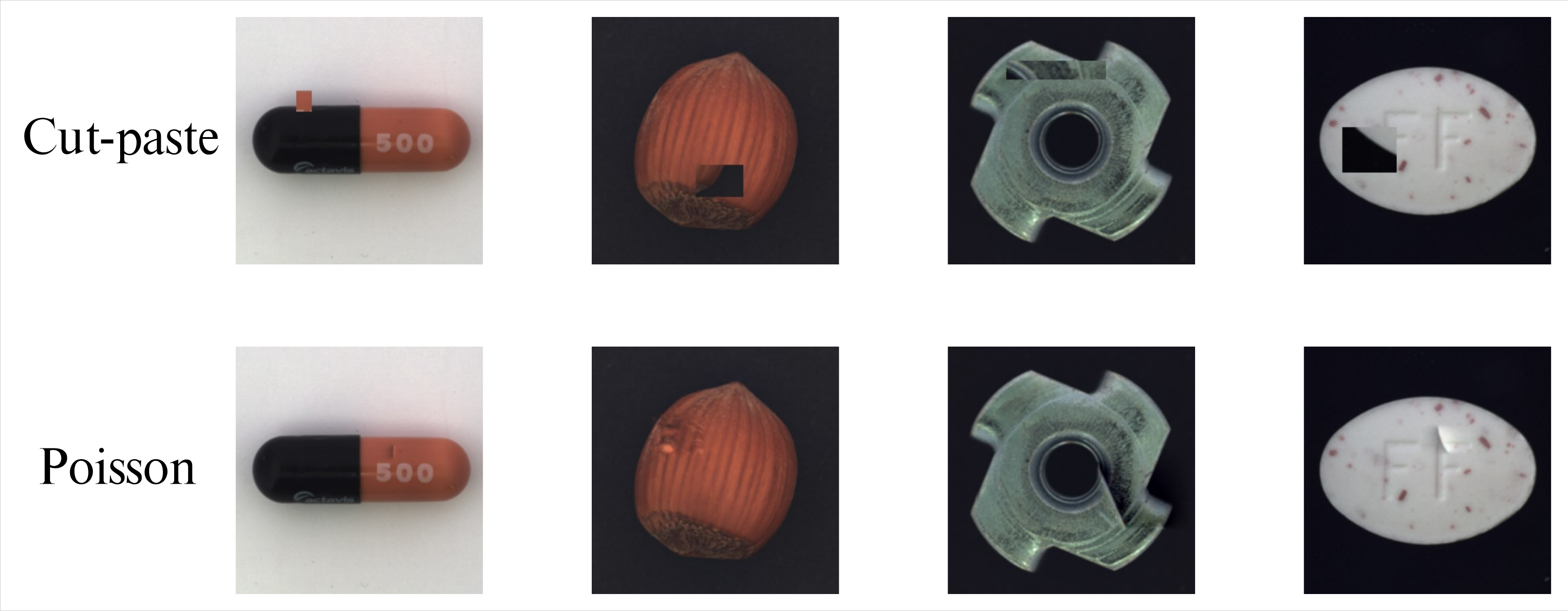} 
    \caption{Illustration of the comparison between cut-paste and poisson image editing. The results of cut-paste exhibit evident discontinuities and the results of poisson image editing are more natural.}
    \label{fig:poisson}
\end{figure}

\noindent\textbf{Anomaly Simulation}
We primarily adopt the approach proposed by NSA~\cite{schluter2022natural} to simulate anomalous data. The NSA~\cite{schluter2022natural} method builds upon the Cut-paste~\cite{li2021cutpaste} technique by incorporating the Poisson image editing~\cite{perez2003poisson} method to alleviate the discontinuity introduced by pasting image segments. Cut-paste~\cite{li2021cutpaste} is a common technique in IAD domain for generating simulated anomaly images. This method involves randomly cropping a block region from an image and then pasting it onto a random location in another image, thus creating a simulated anomalous portion. Simulated anomaly samples can significantly enhance the performance of IAD models, but this procedure often results in noticeable discontinuities, as illustrated in Figure~\ref{fig:poisson}. The Poisson editing method~\cite{perez2003poisson} has been developed to seamlessly clone an object from one image into another image by solving the Poisson partial differential equations. 

\begin{figure}[h]
    \centering
    \includegraphics[width=0.5\columnwidth]{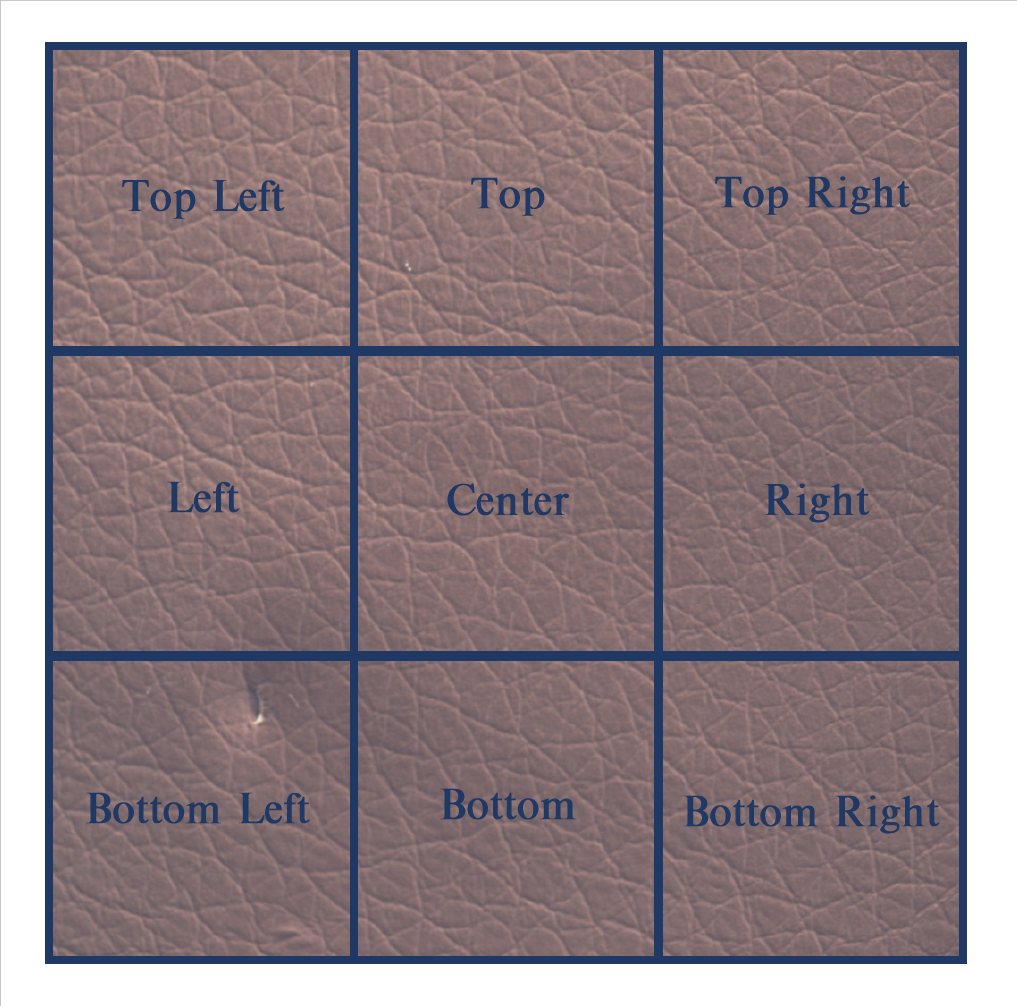} 
    \caption{Illustration of the $3\times 3$ grid of image, which is used to let LLM verbally indicate the abnormal position.}
    \label{fig:3x3}
\end{figure}

\noindent\textbf{Question and Answer Content}
To conduct prompt tuning on the LVLM, we generate corresponding textual queries based on the simulated anomalous images. Specifically, each query consists of two components. The first part involves a description of the input image, providing information about the objects present in the image and their expected attributes, such as \textit{This is a photo of leather, which should be brown and without any damage, flaw, defect, scratch, hole or broken part.} The second part queries the presence of anomalies within the object, namely \textit{Is there any anomaly in the image?} The LVLM firstly responds to whether anomalies are present. If anomalies are detected, the model continues to specify the number and location of the anomalous areas, such as \textit{Yes, there is an anomaly in the image, at the bottom left of the image.} or \textit{No, there are no anomalies in the image.} We divide the image into a grid of $3 \times 3$ distinct regions to facilitate the LVLM in verbally indicating the positions of anomalies, as shown in Figure~\ref{fig:3x3}. The descriptive content about the image furnishes the LVLM with foundational knowledge of the input image, aiding in the model's better comprehension of the image contents. However, during practical applications, users may opt to omit this descriptive input, and the model is still capable of performing IAD task based solely on the provided image input. Detailed description for each category are provided in Appendix C.

Prompts fed to the LLM typically follow the format:

\textit{\#\#\# Human: $<$Img$>$$E_{img}$$<$/Img$>$$E_{prompt}$[Image Description]Is there any anomaly in the image?\#\#\#Assistant:}

$E_{img} \in \mathbb{R}^{C_{emb}}$ represents the image embedding being processed through the image encoder and linear layer, $E_{prompt} \in \mathbb{R}^{(n_1+n_2)\times C_{emb}}$ refers to the prompt embeddings generated by the prompt learner, and \textit{[Image Description]} corresponds to the textual description of the image.

\subsection{Loss Functions}
To train the decoder and prompt learner, we primarily employed three loss functions: cross-entropy loss, focal loss~\cite{lin2017focal}, and dice loss~\cite{milletari2016v}. The latter two are primarily utilized to enhance the pixel-level localization accuracy of the decoder.

\begin{table*}[]
  \centering
  \begin{tabular}{@{}cccccccc@{}}
  \toprule
  \multirow{2}{*}{Setup} &
    \multirow{2}{*}{Method} &
    \multicolumn{3}{c}{MVTec-AD} &
    \multicolumn{3}{c}{VisA} \\ \cmidrule(l){3-8} 
   &
     &
    Image-AUC &
    Pixel-AUC &
    Accuracy &
    Image-AUC &
    Pixel-AUC &
    Accuracy \\ \midrule
  \multirow{5}{*}{1-shot} &
    SPADE &
    81.0 $\pm$ 2.0 &
    91.2 $\pm$ 0.4 &
    - &
    79.5 $\pm$ 4.0 &
    95.6 $\pm$ 0.4 &
    - \\
   &
    PaDiM &
    76.6 $\pm$ 3.1 &
    89.3 $\pm$ 0.9 &
    - &
    62.8 $\pm$ 5.4 &
    89.9 $\pm$ 0.8 &
    - \\
   &
    PatchCore &
    83.4 $\pm$ 3.0 &
    92.0 $\pm$ 1.0 &
    - &
    79.9 $\pm$ 2.9 &
    95.4 $\pm$ 0.6 &
    - \\
   &
    WinCLIP &
    93.1 $\pm$ 2.0 &
    95.2 $\pm$ 0.5 &
    - &
    83.8 $\pm$ 4.0 &
    \textbf{96.4 $\pm$ 0.4} &
    - \\ \cmidrule(l){2-8} 
   &
    \textbf{AnomalyGPT~(ours)} &
    \textbf{94.1 $\pm$ 1.1} &
    \textbf{95.3 $\pm$ 0.1} &
    \textbf{86.1 $\pm$ 1.1} &
    \textbf{87.4 $\pm$ 0.8} &
    96.2 $\pm$ 0.1 &
    \textbf{77.4 $\pm$ 1.0} \\ \midrule
  \multirow{5}{*}{2-shot} &
    SPADE &
    82.9 $\pm$ 2.6 &
    92.0 $\pm$ 0.3 &
    - &
    80.7 $\pm$ 5.0 &
    96.2 $\pm$ 0.4 &
    - \\
   &
    PaDiM &
    78.9 $\pm$ 3.1 &
    91.3 $\pm$ 0.7 &
    - &
    67.4 $\pm$ 5.1 &
    92.0 $\pm$ 0.7 &
    - \\
   &
    PatchCore &
    86.3 $\pm$ 3.3 &
    93.3 $\pm$ 0.6 &
    - &
    81.6 $\pm$ 4.0 &
    96.1 $\pm$ 0.5 &
    - \\
   &
    WinCLIP &
    94.4 $\pm$ 1.3 &
    \textbf{96.0 $\pm$ 0.3} &
    - &
    84.6 $\pm$ 2.4 &
    \textbf{96.8 $\pm$ 0.3} &
    - \\ \cmidrule(l){2-8} 
   &
    \textbf{AnomalyGPT~(ours)} &
    \textbf{95.5 $\pm$ 0.8} &
    95.6 $\pm$ 0.2&
    \textbf{84.8 $\pm$ 0.8} &
    \textbf{88.6 $\pm$ 0.7} &
    96.4 $\pm$ 0.1&
    \textbf{77.5 $\pm$ 0.3} \\ \midrule
  \multirow{5}{*}{4-shot} &
    SPADE &
    84.8 $\pm$ 2.5 &
    92.7 $\pm$ 0.3 &
    - &
    81.7 $\pm$ 3.4 &
    96.6 $\pm$ 0.3 &
    - \\
   &
    PaDiM &
    80.4 $\pm$ 2.5 &
    92.6 $\pm$ 0.7 &
    - &
    72.8 $\pm$ 2.9 &
    93.2 $\pm$ 0.5 &
    - \\
   &
    PatchCore &
    88.8 $\pm$ 2.6 &
    94.3 $\pm$ 0.5 &
    - &
    85.3 $\pm$ 2.1 &
    96.8 $\pm$ 0.3 &
    - \\
   &
    WinCLIP &
    95.2 $\pm$ 1.3 &
    96.2 $\pm$ 0.3 &
    - &
    87.3 $\pm$ 1.8 &
    \textbf{97.2 $\pm$ 0.2} &
    - \\ \cmidrule(l){2-8} 
   &
    \textbf{AnomalyGPT~(ours)} &
    \textbf{96.3 $\pm$ 0.3} &
    \textbf{96.2 $\pm$ 0.1} &
    \textbf{85.0 $\pm$ 0.3} &
    \textbf{90.6 $\pm$ 0.7} &
    96.7 $\pm$ 0.1 &
    \textbf{77.7 $\pm$ 0.4} \\ \bottomrule
  \end{tabular}
  \caption{Few-shot IAD results on MVTec-AD and VisA datasets. Results are listed as the average of 5 runs and the best-performing method is in \textbf{bold}. The results for SPADE, PaDiM, PatchCore and WinCLIP are reported from~\cite{jeong2023winclip}.}
  \label{tab:few-shot}
  \end{table*}

\begin{table}[]
  \centering
  \small
  \begin{tabular}{@{}cccc@{}}
  \toprule
  Method              & Image-AUC     & Pixel-AUC     & Accuracy     \\ \midrule
  PaDiM~(Unified)               & 84.2          & 89.5          & -             \\
  JNLD~(Unified)               & 91.3          & 88.6          & -             \\
  UniAD               & 96.5          & \textbf{96.8} & -             \\
  \textbf{AnomalyGPT~(ours)} & \textbf{97.4} & 93.1          & \textbf{93.3} \\ \bottomrule
  \end{tabular}
  \caption{Unsupervised anomaly detection results on MVTec-AD dataset. The best-performing method is in \textbf{bold} and the results for PaDiM and JNLD are reported from~\cite{zhao2023omnial}.}
  \label{tab:unified}
  \end{table}

\noindent\textbf{Cross-Entropy Loss}
Cross-entropy loss is commonly employed for training language models, which quantifies the disparity between the text sequence generated by the model and the target text sequence. The formula is as follows:
\begin{equation}
    L_{ce} = -\sum_{i=1}^{n}{y_i log(p_i)},\label{eq:ce}
\end{equation}
where $n$ is the number of tokens,  $y_i$ is the true label for token $i$ and $p_i$ is the predicted probability for token $i$.

\noindent\textbf{Focal Loss}
Focal loss~\cite{lin2017focal} is commonly used in object detection and semantic segmentation to address the issue of class imbalance, which introduces an adjustable parameter $\gamma$ to modify the weight distribution of cross-entropy loss, emphasizing samples that are difficult to classify. In IAD task, where most regions in anomaly images are still normal, employing focal loss can mitigate the problem of class imbalance. Focal loss can be calculated by Eq.~(\ref{eq:focal}):
\begin{equation}
    L_{focal} = -\frac{1}{n}\sum_{i=1}^{n}{(1-p_i)^{\gamma}log(p_i)},\label{eq:focal}
\end{equation}
where $n = H \times W$ represents the total number of pixels, $p_i$ is the predicted probability of the positive classes and $\gamma$ is a tunable parameter for adjusting the weight of hard-to-classify samples. In our implementation, we set $\gamma$ to 2.

\noindent\textbf{Dice Loss}
Dice loss~\cite{milletari2016v} is a commonly employed loss function in semantic segmentation tasks. It is based on the dice coefficient and can be calculated by Eq.~(\ref{eq:dice}):
\begin{equation}
    L_{dice} = -\frac{\sum_{i=1}^{n}{y_i\hat{y}_i}}{\sum_{i=1}^{n}{y_i^2}+\sum_{i=1}^{n}{\hat{y}^2_i}},\label{eq:dice}
\end{equation}
where $n = H \times W$, $y_i$ is the output of decoder and $\hat{y}_i$ is the ground truth value.

Finally, the overall loss function is defined as:
\begin{equation}
    L = \alpha L_{ce}+\beta L_{focal}+\delta L_{dice},\label{eq:loss}
\end{equation}
where $\alpha, \beta, \delta$ are coefficients to balance the three loss functions, which are set to 1 by default in our experiments.

\begin{table*}[]
  \centering
  \small
  \begin{tabular}{@{}cccccccccc@{}}
      \toprule
      \multirow{2}{*}{Decoder} &
        \multirow{2}{*}{Prompt learner} &
        \multirow{2}{*}{LLM} &
        \multirow{2}{*}{LoRA} &
        \multicolumn{3}{c}{MVTec-AD~(unsupervised)} &
        \multicolumn{3}{c}{VisA~(1-shot)} \\ \cmidrule(l){5-10} 
                   &              &              &              & Image-AUC & Pixel-AUC & Accuracy & Image-AUC & Pixel-AUC & Accuracy \\ \midrule
                   &              & $\checkmark$ &              & -         & -         & 72.2     & -         & -         & 56.5     \\
                   & $\checkmark$ & $\checkmark$ &              & -         & -         & 73.4     & -         & -         & 56.6     \\
                   &              & $\checkmark$ & $\checkmark$ & -         & -         & 79.8     & -         & -         & 63.4     \\
      $\checkmark$ &              & $\checkmark$ &              & 97.1      & 90.9      & 72.2     & 85.8      & 96.2      & 56.5     \\
      $\checkmark$ &              & $\checkmark$ & $\checkmark$ & 97.1      & 90.9      & 84.2     & 85.8      & 96.2      & 64.7     \\
      $\checkmark$ & $\checkmark$ & $\checkmark$ & $\checkmark$ & 96.0      & 88.1      & 83.9     & 85.8      & \textbf{96.5}      & 72.7     \\
      $\checkmark$ &              &              &              & 97.1      & 90.9      & 90.3     & 85.8      & 96.2      & 75.4     \\
      $\checkmark$ &
        $\checkmark$ &
        $\checkmark$ &
         &
        \textbf{97.4} &
        \textbf{93.1} &
        \textbf{93.3} &
        \textbf{87.4} &
        96.2 &
        \textbf{77.4} \\ \bottomrule
      \end{tabular}
  \caption{Results of ablation studies. The $\checkmark$ in ``Decoder'' and ``Prompt learner'' columns indicate module inclusion. The $\checkmark$ in ``LLM'' column denotes whether use LLM for inference and the $\checkmark$ in ``LoRA'' column denotes whether use LoRA to fine-tune LLM. In settings without LLM, the maximum anomaly score from normal samples is used as the classification threshold. In settings without decoder, due to the sole textual output from the LLM, we cannot compute image-level and pixel-level AUC.}
  \label{tab:ablation}
  \end{table*}

\section{Experiments}
\noindent\textbf{Datasets}
We conduct experiments primarily on the MVTec-AD~\cite{bergmann2019mvtec} and VisA~\cite{zou2022spot} datasets. The MVTec-AD dataset comprises 3629 training images and 1725 testing images across 15 different categories, making it one of the most popular datasets for IAD. The training images only consist of normal images, while the testing images contain both normal and anomalous images. The image resolutions vary from 700$\times$700 to 1024$\times$1024. VisA, a newly introduced IAD dataset, contains 9621 normal images and 1200 anomalous images across 12 categories, with resolutions approximately around 1500$\times$1000. Consistent with previous IAD methods, we only use the normal data from these datasets for training.

\noindent\textbf{Evaluation Metrics}
Following existing IAD methods, we employ the Area Under the Receiver Operating Characteristic~(AUC) as our evaluation metric, with image-level and pixel-level AUC used to assess anomaly detection and anomaly localization performance, respectively. However, our proposed approach uniquely allows for determining the presence of anomalies without the need for manually-set thresholds. Therefore, we also utilize the image-level accuracy to evaluate the performance of our method.

\noindent\textbf{Implementation Details}
We utilize ImageBind-Huge~\cite{girdhar2023imagebind} as the image encoder and Vicuna-7B~\cite{chiang2023vicuna} as the inferential LLM, connected through a linear layer. We initialize our model using pre-trained parameters from PandaGPT~\cite{su2023pandagpt}. We set the image resolution at 224$\times$224 and feed the outputs from the 8th, 16th, 24th, and 32nd layers of ImageBind-Huge's image encoder to the image decoder. Training is conducted on two RTX-3090 GPUs over 50 epochs, with a learning rate of 1e-3 and a batch size of 16. Linear warm-up and a one-cycle cosine learning rate decay strategy are applied. We perform alternating training using both the pre-training data of PandaGPT~\cite{su2023pandagpt} and our anomaly image-text data. Only the decoder and prompt learner undergo parameter updates, while the remaining parameters are all kept frozen.

\begin{figure}[t]
    \centering
    \includegraphics[width=0.95\columnwidth]{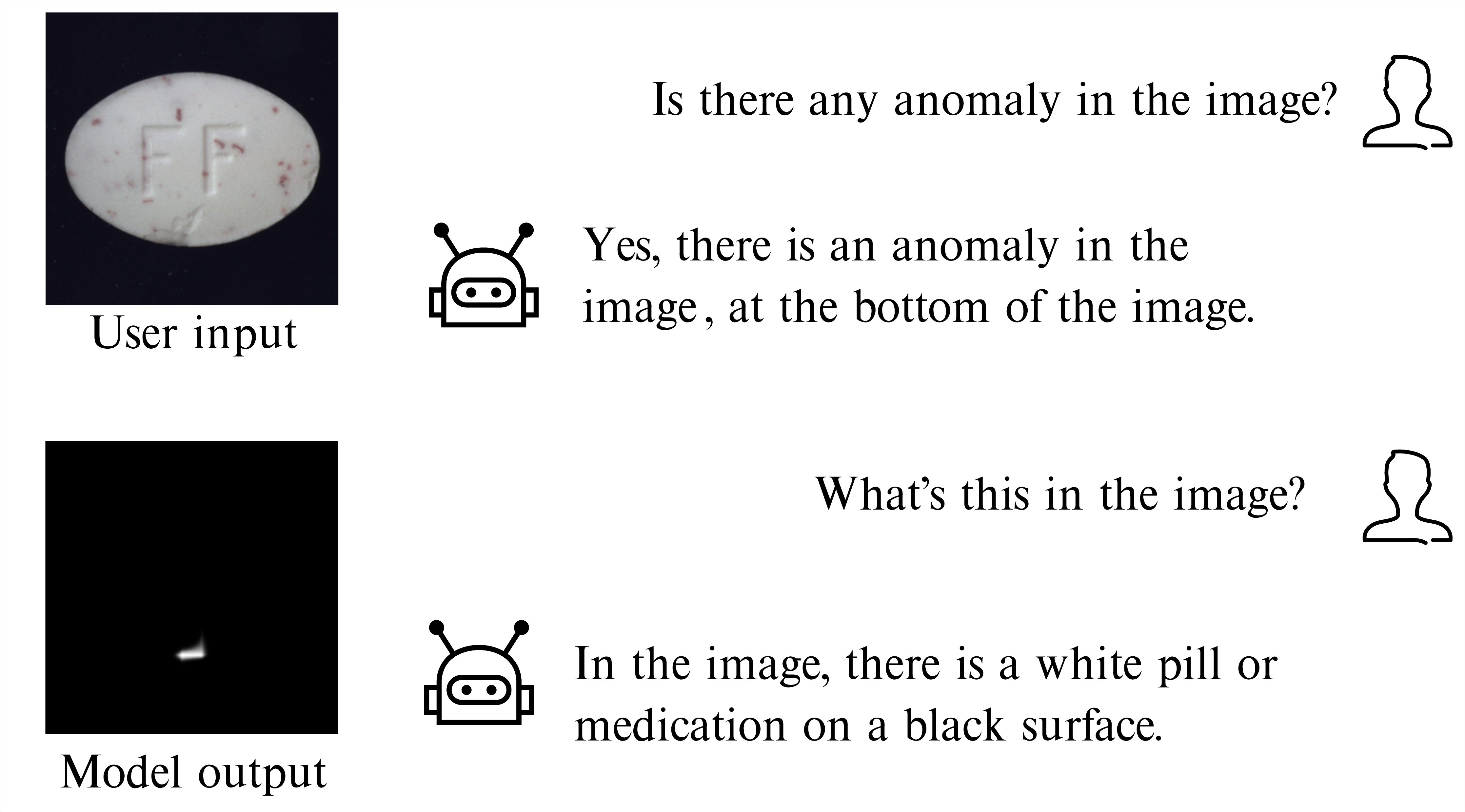} 
    \caption{Qualitative example of AnomalyGPT in the unsupervised setting. AnomalyGPT is capable of detecting anomaly, pinpointing its location, providing pixel-level localization results and answering questions about the image.}
    \label{fig:case_1}
\end{figure}

\subsection{Quantitative Results}

\noindent\textbf{Few-Shot Industrial Anomaly Detection}
We compare our work with prior few-shot IAD methods, selecting SPADE~\cite{cohen2020sub}, PaDiM~\cite{defard2021padim}, PatchCore~\cite{roth2022towards}, and WinCLIP~\cite{jeong2023winclip} as the baselines. The results are presented in Table~\ref{tab:few-shot}. Across both datasets, our method notably outperforms previous approaches in terms of image-level AUC and achieves competitive pixel-level AUC and good accuracy.

\noindent\textbf{Unsupervised Industrial Anomaly Detection}
In the setting of unsupervised training with a large number of normal samples, given that our method trains a single model on samples from all classes within a dataset, we selected UniAD~\cite{you2022unified}, which is trained under the same setup, as a baseline for comparison. Additionally, we compare our model with PaDiM~\cite{defard2021padim} and JNLD~\cite{zhao2022just} using the same unified setting. The results on MVTec-AD dataset are presented in Table~\ref{tab:unified}.

\begin{figure}[t]
    \centering
    \includegraphics[width=0.95\columnwidth]{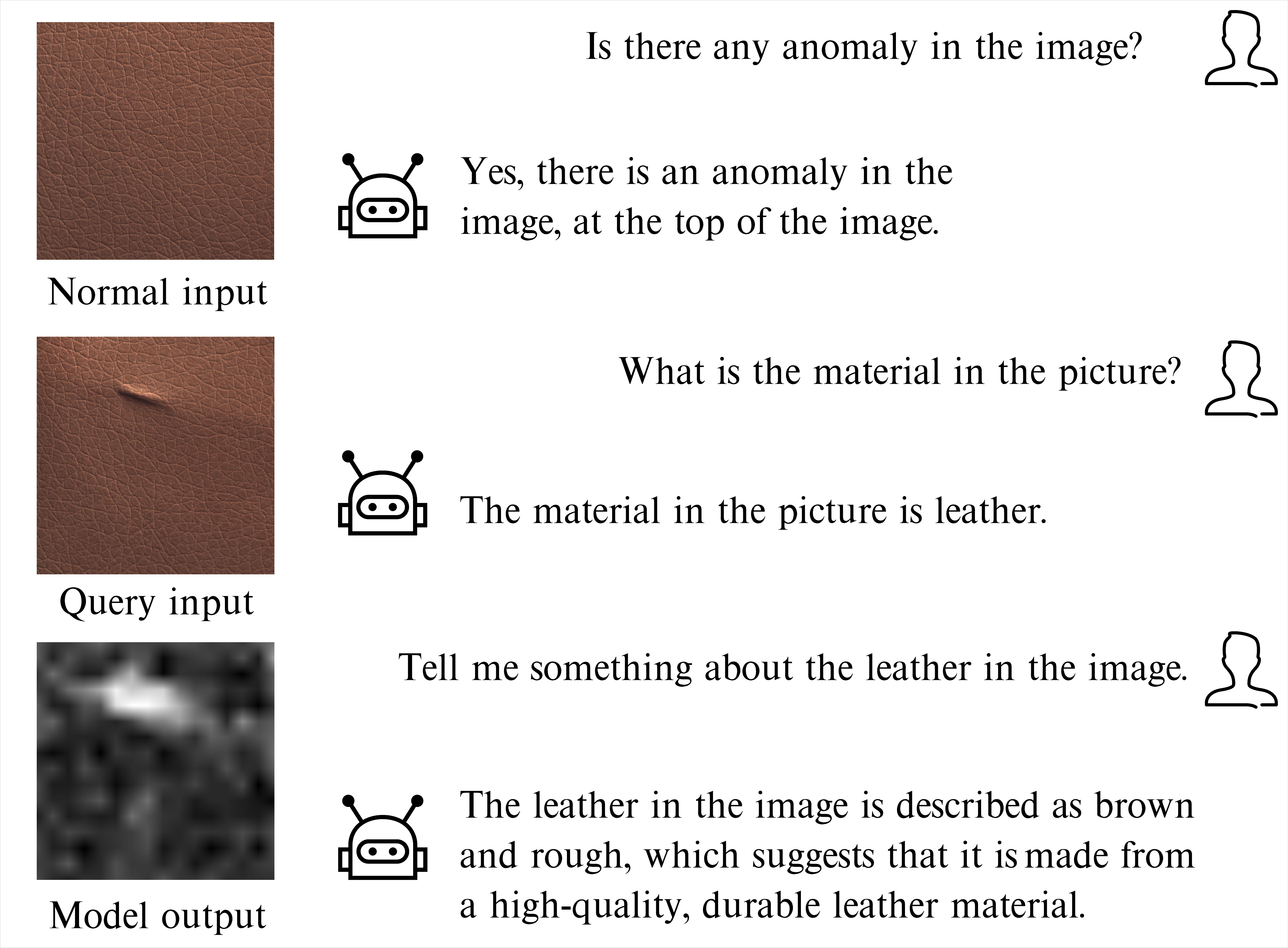} 
    \caption{Qualitative example of AnomalyGPT in the \textbf{one-normal-shot} setting. The localization performance is slightly lower compared to the unsupervised setting due to the absence of parameter training.}
    \label{fig:case_2}
\end{figure}

\subsection{Qualitative Examples}
Figure~\ref{fig:case_1} illustrates the performance of our AnomalyGPT in unsupervised anomaly detection, and Figure~\ref{fig:case_2} showcases the results in the 1-shot in-context learning. Our model is capable of indicating the presence of anomalies, pinpointing their locations, and providing pixel-level localization results. Users can engage in multi-turn dialogues related to image content. In the 1-shot in-context learning setting, due to the absence of training, the model's localization performance is slightly lower than the unsupervised setting. More qualitative examples can be found in Appendix D.

\subsection{Ablation Studies}
To prove the efficacy of each proposed module, extensive ablation experiments are conducted on both the MVTec-AD and VisA datasets. We primarily focus on four aspects: the decoder, prompt learner, the usage of LLM for inference, and the utilization of LoRA to fine-tune the LLM. The principal results are presented in Table~\ref{tab:ablation}. Unsupervised training and testing are carried out on the MVTec-AD dataset, while the one-shot performance is evaluated on the visa dataset. It can be observed that the decoder demonstrates impressive pixel-level anomaly localization performance. Compared to manually-set thresholds, the LLM exhibits superior inference accuracy and provides additional functionality. Furthermore, prompt tuning outperforms LoRA in terms of accuracy and transferability.

\section{Conclusion}
We introduce AnomalyGPT, a novel conversational IAD vision-language model, leveraging the powerful capabilities of LVLM. AnomalyGPT can determine whether an image contains anomalies and pinpoint their locations without the need for manually specified thresholds. Furthermore, AnomalyGPT enables multi-turn dialogues focused on anomaly detection and demonstrates remarkable performance in few-shot in-context learning. The effectiveness of AnomalyGPT is validated on two common datasets. Our work delves into the potential application of large visual language models in anomaly detection, offering fresh ideas and possibilities for the field of industrial anomaly detection.

{
    \small
    \bibliographystyle{ieeenat_fullname}
    \bibliography{main}
}

\appendix

\onecolumn

\begin{center}
    {\bf {\LARGE Supplementary Material}}
\end{center}
\begin{center}
{\bf {\large AnomalyGPT: Detecting Industrial Anomalies Using Large Vision-Language Models \\[20pt]}}
\end{center}

\section{More Experimental Results of Existing IAD Methods}

As described in the paper, existing IAD methods solely provide anomaly scores for test samples. However, anomaly scores alone do not allow users to determine the presence of anomalies because they don't know the threshold to distinguish between normal and abnormal samples. The threshold varies considerably for each category of objects. Only by concurrently obtaining both normal and anomalous samples for each object category, along with their respective anomaly scores, can users identify the optimal classification threshold, which contradicts the original intent of IAD task that only utilize normal data.

One potential solution to this problem involves utilizing the maximum anomaly score from the normal samples of each category in the training set as the threshold. However, this approach is solely suited for ``one-class-one-model" methods, which are designed specifically for detecting objects of a certain category within a specific environment. When presented with a previously unseen test sample, the model remains uncertain about which threshold to apply for decision-making. Thus, a unified threshold applicable across all categories is needed, which is challenging for existing IAD techniques.

\begin{figure}[h]
  \centering
  \includegraphics[width=0.95\columnwidth]{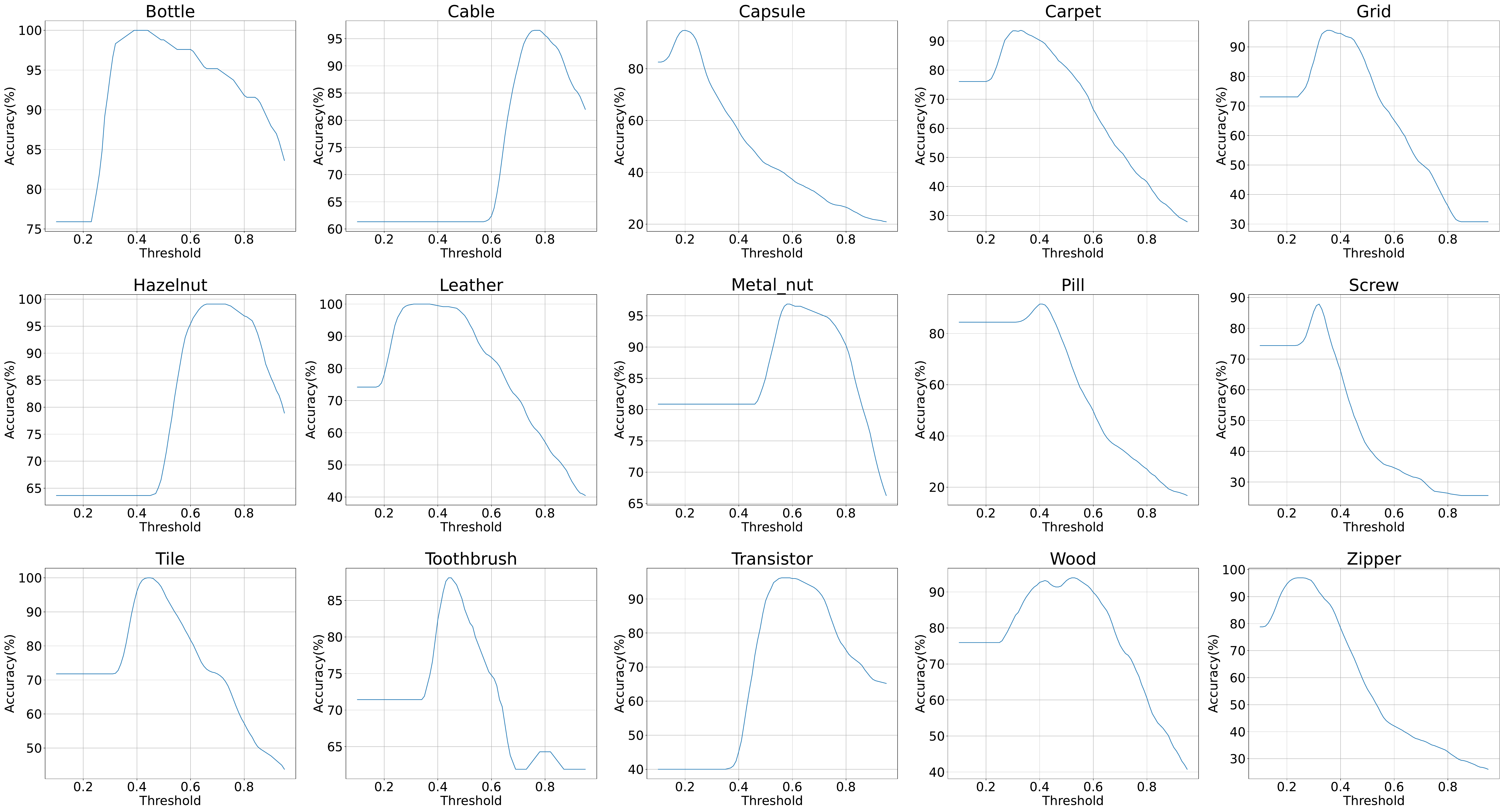} 
  \caption{Experimental results of PatchCore~\cite{roth2022towards} on the MVTec-AD~\cite{bergmann2019mvtec} dataset across each category under different thresholds. The optimal threshold varies considerably for each category of objects.}
  \label{fig:threshold_patchcore}
\end{figure}

\begin{figure}[h]
  \centering
  \includegraphics[width=0.95\columnwidth]{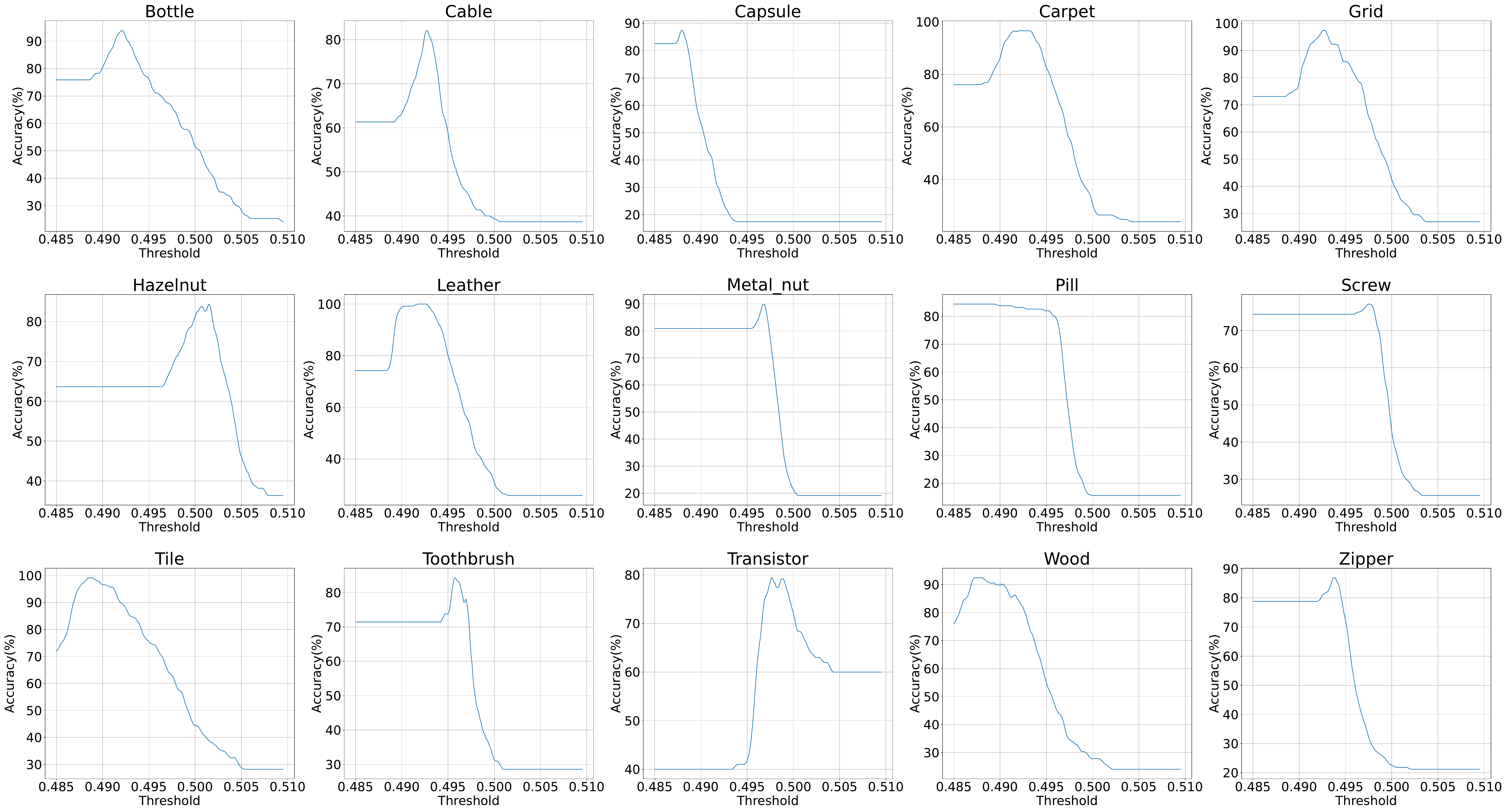} 
  \caption{Experimental results of WinCLIP~\cite{jeong2023winclip} on the MVTec-AD~\cite{bergmann2019mvtec} dataset across each category under different thresholds. The optimal threshold varies considerably for each category of objects.}
  \label{fig:threshold_winclip}
\end{figure}

We conduct experiments on two representative IAD methods, PatchCore~\cite{roth2022towards} and WinCLIP~\cite{jeong2023winclip}. PatchCore~\cite{roth2022towards} achieves an Image-level AUC of 99.3\% on the MVTec-AD dataset, while WinCLIP~\cite{jeong2023winclip} is the state-of-the-art method for few-shot IAD. We assess the accuracy of both methods across individual categories at varying thresholds. It can be  observed that the threshold exerts a significant influence on the performance of these two methods. Furthermore, a singular threshold displays markedly different efficacies across disparate categories. Hence, it becomes challenging to ascertain an optimal threshold unless experimental trials are conducted on test sets containing anomalous samples for each category. Figures~\ref{fig:threshold_patchcore} and Figure~\ref{fig:threshold_winclip} delineate the outcomes of PatchCore~\cite{roth2022towards} and WinCLIP~\cite{jeong2023winclip} on the MVTec-AD~\cite{bergmann2019mvtec} dataset across each category under different threshold settings.

\clearpage

\section{Normal and Abnormal Texts}

Following WinCLIP~\cite{jeong2023winclip}, we utilize the compositional prompt ensemble to obtain texts presenting normality and abnormality. Specifically, we consider two levels of texts: (a) state-level, and (b) template level. The complete text can be composed by replacing the token \verb|[c]| in a template-level text with one of state-level text and replacing the token \verb|[o]| with the object's name. When the item's name is unavailable, the term ``object" is adopted as the name for the item. Table~\ref{tab:comp_prompt} provides a detailed list of the multi-level texts.

\begin{table}[!h]
  \noindent\begin{minipage}[t]{0.32\linewidth}
    (a) \emph{State}-level (normal)
    
    {\tt \small
    \begin{itemize}
        \item c := "[o]"
        \item c := "flawless [o]"
        \item c := "perfect [o]"
        \item c := "unblemished [o]"
        \item c := "[o] without flaw"
        \item c := "[o] without defect"
        \item c := "[o] without damage"
    \end{itemize}
    }
    
    \emph{State}-level (anomaly)
    
    {\tt \small
    \begin{itemize}
        \item c := "damaged [o]"
        \item c := "broken [o]"
        \item c := "[o] with flaw"
        \item c := "[o] with defect"
        \item c := "[o] with damage"
    \end{itemize}
    }
    \end{minipage}
    \hfill
    \begin{minipage}[t]{0.33\linewidth}
    (b) \emph{Template}-level
    
    {\tt \small
    \begin{itemize}
    \item "a cropped photo of the [c]."
    \item "a cropped photo of a [c]."
    \item "a close-up photo of a [c]."
    \item "a close-up photo of the [c]."
    \item "a bright photo of a [c]." 
    \item "a bright photo of the [c]."
    \item "a dark photo of the [c]."
    \item "a dark photo of a [c]."
    \item "a jpeg corrupted photo of a [c]."
    \item "a jpeg corrupted photo of the [c]."

    \end{itemize}
    }
    \end{minipage}
    \begin{minipage}[t]{0.33\linewidth}
    {\tt \small
    \begin{itemize}
    
    \item "a blurry photo of the [c]."
    \item "a blurry photo of a [c]."
    \item "a photo of a [c]."
    \item "a photo of the [c]."
    \item "a photo of a small [c]."
    \item "a photo of the small [c]."
    \item "a photo of a large [c]."
    \item "a photo of the large [c]."
    \item "a photo of the [c] for visual inspection."
    \item "a photo of a [c] for visual inspection."
    \item "a photo of the [c] for anomaly detection."
    \item "a photo of a [c] for anomaly detection."
    \end{itemize}
    }
    \end{minipage}
    
    \caption{Lists of multi-level texts considered in this paper to present normal and abnormal semantics.}
    \label{tab:comp_prompt}
    \end{table}

\clearpage
\section{Detailed Image Description}

As mentioned in the paper, prompts fed to the LLM typically follow the format: \\[3pt]
\indent\textit{\#\#\# Human: $<$Img$>$ $E_{img}$ $<$/Img$>$ $E_{prompt}$ [Image Description] Is there any anomaly in the image? \#\#\# Assistant:} \\[3pt]
\indent The \textit{[Image Description]} part involves a description of the input image, providing information about the objects present in the image and their expected attributes. Such description furnishes the LVLM with foundational knowledge of the input image, aiding in the model's better comprehension of the image contents. The detailed description of every category in MVTec-AD~\cite{bergmann2019mvtec} and VisA~\cite{zou2022spot} datasets can be found in Table~\ref{tab:mvtec-description} and Table~\ref{tab:visa-description}. Note that users can omit this descriptive input, and the model is still capable of performing IAD task based solely on the provided image input.

\begin{table}[!h]
    \centering
    \begin{tabular}{@{}cc@{}}
      \toprule
      \textbf{Class} &
        \textbf{Image description} \\ \midrule
      Bottle &
        \begin{tabular}[c]{@{}c@{}}This is a photo of a bottle for anomaly detection, which should be round and without any \\ damage, flaw, defect, scratch, hole or broken part.\end{tabular} \\ \midrule
      Cable &
        \begin{tabular}[c]{@{}c@{}}This is a photo of three cables for anomaly detection, they are green, blue and grey, which cannot \\ be missed or swapped and should be without any damage, flaw, defect, scratch, hole or broken part.\end{tabular} \\ \midrule
      Capsule &
        \begin{tabular}[c]{@{}c@{}}This is a photo of a capsule for anomaly detection, which should be black and orange, \\ with print `500' and without any damage, flaw, defect, scratch, hole or broken part.\end{tabular} \\ \midrule
      Carpet &
        \begin{tabular}[c]{@{}c@{}}This is a photo of carpet for anomaly detection, which should be without any \\ damage, flaw, defect, scratch, hole or broken part.\end{tabular} \\ \midrule
      Grid &
        \begin{tabular}[c]{@{}c@{}}This is a photo of grid for anomaly detection, which should be without any \\ damage, flaw, defect, scratch, hole or broken part.\end{tabular} \\ \midrule
      Hazelnut &
        \begin{tabular}[c]{@{}c@{}}This is a photo of a hazelnut for anomaly detection, which should be without any \\ damage, flaw, defect, scratch, hole or broken part.\end{tabular} \\ \midrule
      Leather &
        \begin{tabular}[c]{@{}c@{}}This is a photo of leather for anomaly detection, which should be brown with patterns and \\ without any  damage, flaw, defect, scratch, hole or broken part.\end{tabular} \\ \midrule
      Metal nut &
        \begin{tabular}[c]{@{}c@{}}This is a photo of a metal nut for anomaly detection, which should be without any \\ damage, flaw, defect, scratch, hole or broken part, and shouldn't be fliped.\end{tabular} \\ \midrule
      Pill &
        \begin{tabular}[c]{@{}c@{}}This is a photo of a pill for anomaly detection, which should be white, with print `FF' \\ and red patterns and without any damage, flaw, defect, scratch, hole or broken part.\end{tabular} \\ \midrule
      Screw &
        \begin{tabular}[c]{@{}c@{}}This is a photo of a screw for anomaly detection, whose tail should be sharp, \\ and without any damage, flaw, defect, scratch, hole or broken part.\end{tabular} \\ \midrule
      Tile &
        \begin{tabular}[c]{@{}c@{}}This is a photo of tile for anomaly detection, which should be without any \\ damage, flaw, defect, scratch, hole or broken part.\end{tabular} \\ \midrule
      Toothbrush &
        \begin{tabular}[c]{@{}c@{}}This is a photo of a toothbrush for anomaly detection, which should be without any \\ damage, flaw, defect, scratch, hole or broken part.\end{tabular} \\ \midrule
      Transistor &
        \begin{tabular}[c]{@{}c@{}}This is a photo of a transistor for anomaly detection, which should be without any \\ damage, flaw, defect, scratch, hole or broken part.\end{tabular} \\ \midrule
      Wood &
        \begin{tabular}[c]{@{}c@{}}This is a photo of wood for anomaly detection, which should be brown with patterns  \\ and without any damage, flaw, defect, scratch, hole or broken part.\end{tabular} \\ \midrule
      Zipper &
        \begin{tabular}[c]{@{}c@{}}This is a photo of a zipper for anomaly detection, which should be without \\ any damage, flaw, defect, scratch, hole or broken part.\end{tabular} \\ \bottomrule
      \end{tabular}

    \caption{Detailed image description for every category in MVTec-AD dataset. The description will be added to the prompts of the corresponding category during training to provide foundational knowledge of the input image.}
    \label{tab:mvtec-description}
    \end{table}
\clearpage

\begin{table}[!h]
  \centering
  \begin{tabular}{@{}cc@{}}
    \toprule
    \textbf{Class} &
      \textbf{Image description} \\ \midrule
    Candle &
      \begin{tabular}[c]{@{}c@{}}This is a photo of 4 candles for anomaly detection, every candle should be round, \\ without any damage, flaw, defect, scratch, hole or broken part.\end{tabular} \\ \midrule
    Capsules &
      \begin{tabular}[c]{@{}c@{}}This is a photo of many small capsules for anomaly detection, every capsule is green and should be \\without any damage, flaw, defect, scratch, hole or broken part.\end{tabular} \\ \midrule
    Cashew &
      \begin{tabular}[c]{@{}c@{}}This is a photo of a cashew for anomaly detection, which should be without any \\ damage, flaw, defect, scratch, hole or broken part.\end{tabular} \\ \midrule
    Chewinggum &
      \begin{tabular}[c]{@{}c@{}}This is a photo of a chewinggum for anomaly detection, which should be white, \\ without any damage, flaw, defect, scratch, hole or broken part.\end{tabular} \\ \midrule
    Fryum &
      \begin{tabular}[c]{@{}c@{}}This is a photo of a fryum for anomaly detection on green background, which should be \\without any damage, flaw, defect, scratch, hole or broken part.\end{tabular} \\ \midrule
    Macaroni1 &
      \begin{tabular}[c]{@{}c@{}}This is a photo of 4 macaronis for anomaly detection, which should be without any \\ damage, flaw, defect, scratch, hole or broken part.\end{tabular} \\ \midrule
    Macaroni2 &
      \begin{tabular}[c]{@{}c@{}}This is a photo of 4 macaronis for anomaly detection, which should be without any \\ damage, flaw, defect, scratch, hole or broken part.\end{tabular} \\ \midrule
    PCB1 &
      \begin{tabular}[c]{@{}c@{}}This is a photo of PCB for anomaly detection, which should be without any \\ damage, flaw, defect, scratch, hole or broken part.\end{tabular} \\ \midrule
    PCB2 &
      \begin{tabular}[c]{@{}c@{}}This is a photo of PCB for anomaly detection, which should be without any \\ damage, flaw, defect, scratch, hole or broken part.\end{tabular} \\ \midrule
    PCB3 &
      \begin{tabular}[c]{@{}c@{}}This is a photo of PCB for anomaly detection, which should be without any \\ damage, flaw, defect, scratch, hole or broken part.\end{tabular} \\ \midrule
    PCB4 &
      \begin{tabular}[c]{@{}c@{}}This is a photo of PCB for anomaly detection, which should be without any \\ damage, flaw, defect, scratch, hole or broken part.\end{tabular} \\ \midrule
   
    Pipe fryum &
      \begin{tabular}[c]{@{}c@{}}This is a photo of a pipe fryum for anomaly detection, which should be without any \\ damage, flaw, defect, scratch, hole or broken part.\end{tabular} \\ \bottomrule
    
    \end{tabular}

  \caption{Detailed image description for every category in VisA dataset. The description will be added to the prompts of the corresponding category during training to provide foundational knowledge of the input image.}
  \label{tab:visa-description}
  \end{table}

\clearpage
\section{More Qualitative Examples}

\begin{figure}[h]
  \centering
  \includegraphics[width=0.7\columnwidth]{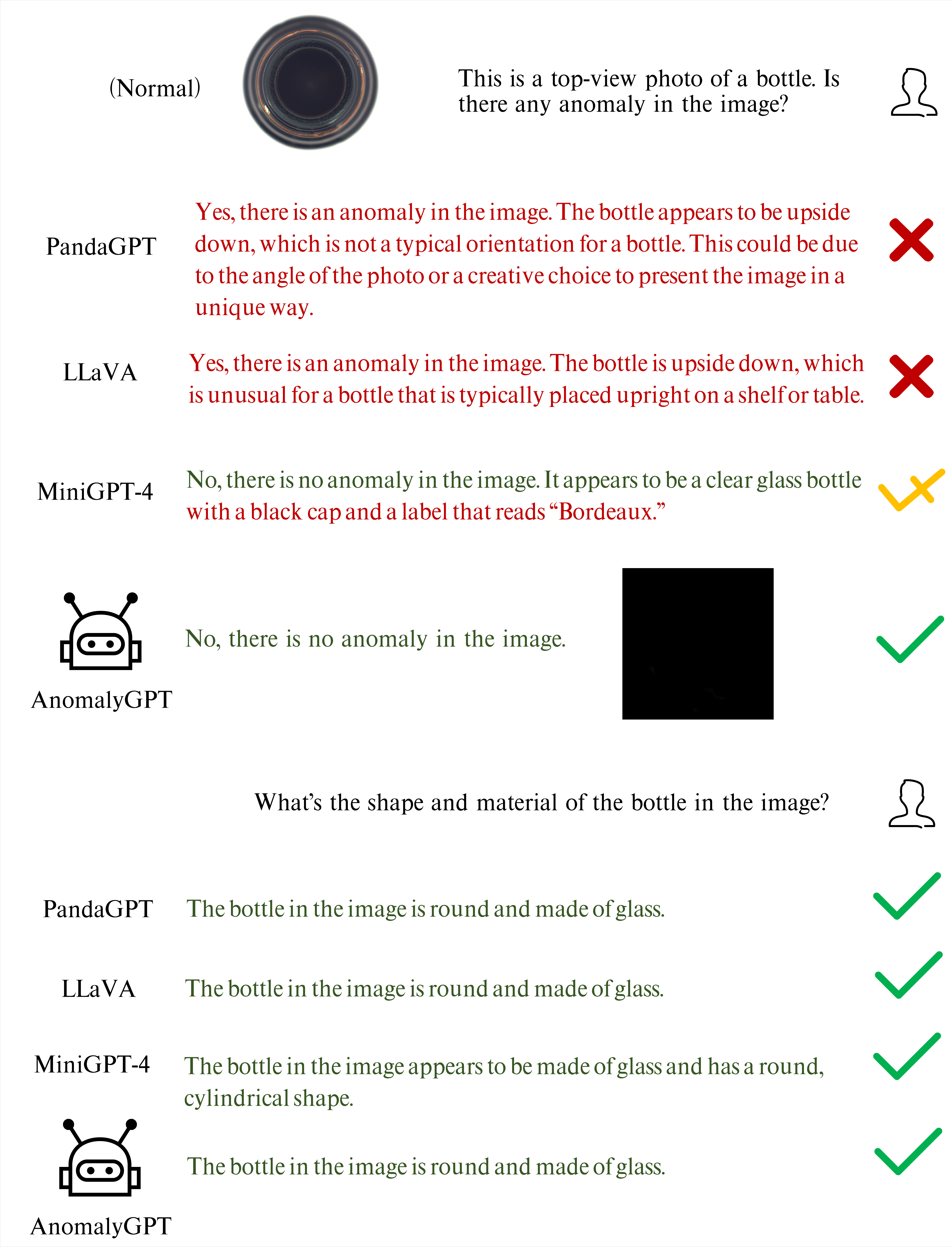} 
  \caption{Comparison between AnomalyGPT, PandaGPT, LLaVA and MiniGPT-4. The input image is a top-view photo of a normal bottle. AnomalyGPT is capable of detecting anomaly, pinpointing its location, providing pixel-level localization results and answering questions about the image.}
  \label{fig:case_c1}
\end{figure}

\begin{figure}[h]
  \centering
  \includegraphics[width=0.7\columnwidth]{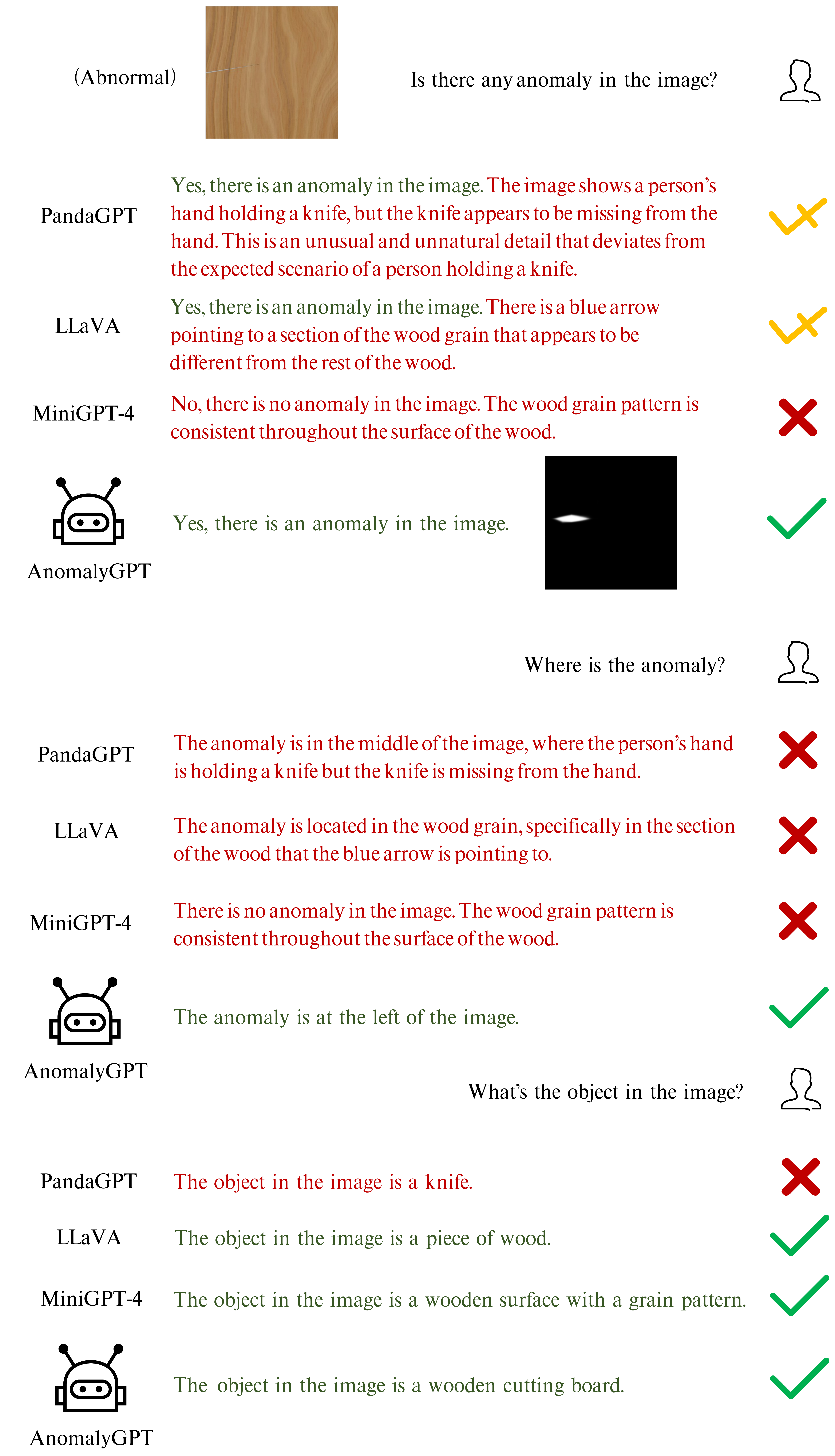} 
  \caption{Comparison between AnomalyGPT, PandaGPT, LLaVA and MiniGPT-4. The input image is a photo of a cutting wood. AnomalyGPT is capable of detecting anomaly, pinpointing its location, providing pixel-level localization results and answering questions about the image.}
  \label{fig:case_c2}
\end{figure}

\begin{figure}[h]
  \centering
  \includegraphics[width=0.75\columnwidth]{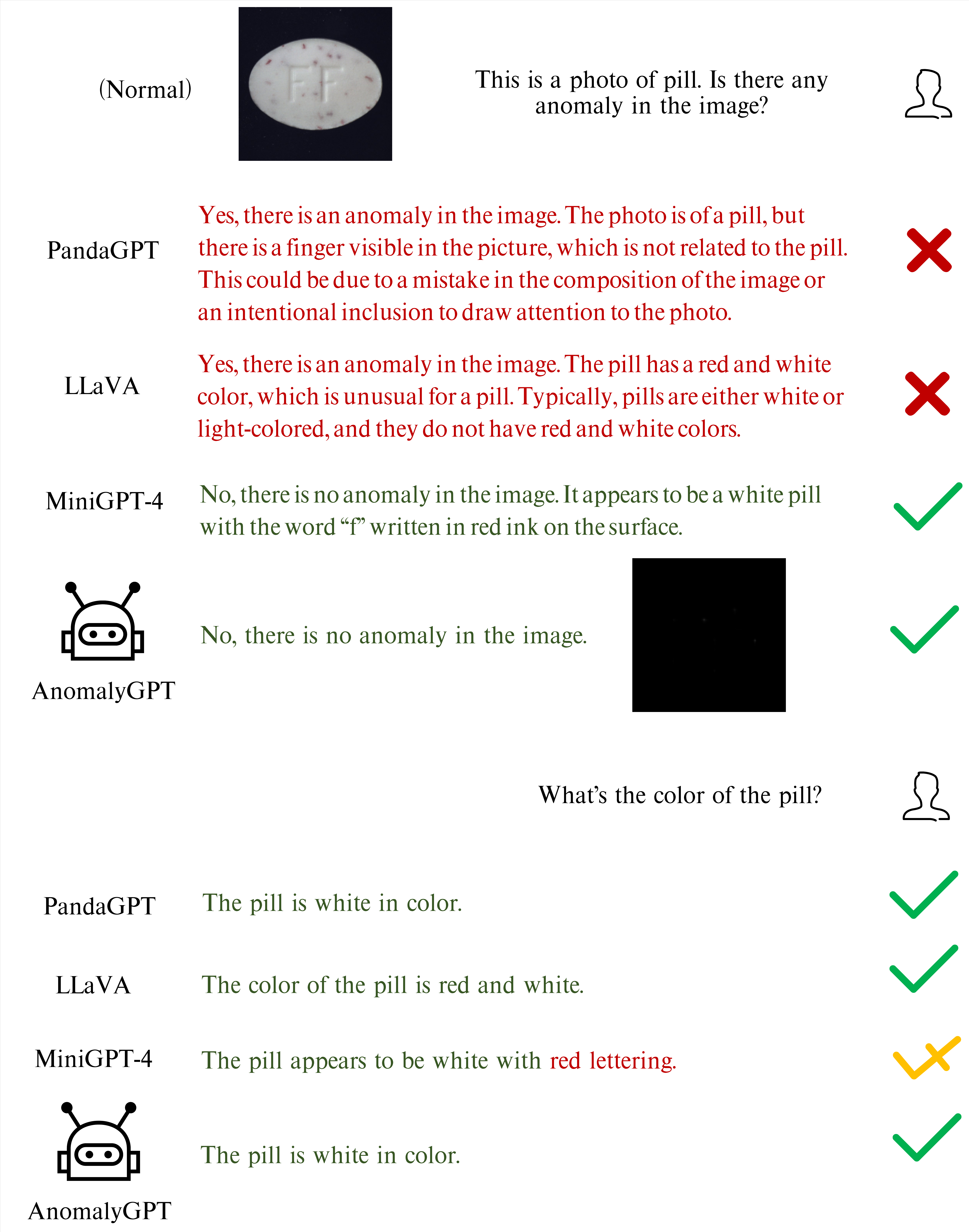} 
  \caption{Comparison between AnomalyGPT, PandaGPT, LLaVA and MiniGPT-4. The input image is a photo of a normal pill. AnomalyGPT is capable of detecting anomaly, pinpointing its location, providing pixel-level localization results and answering questions about the image.}
  \label{fig:case_6}
\end{figure}

\begin{figure}[h]
  \centering
  \includegraphics[width=0.7\columnwidth]{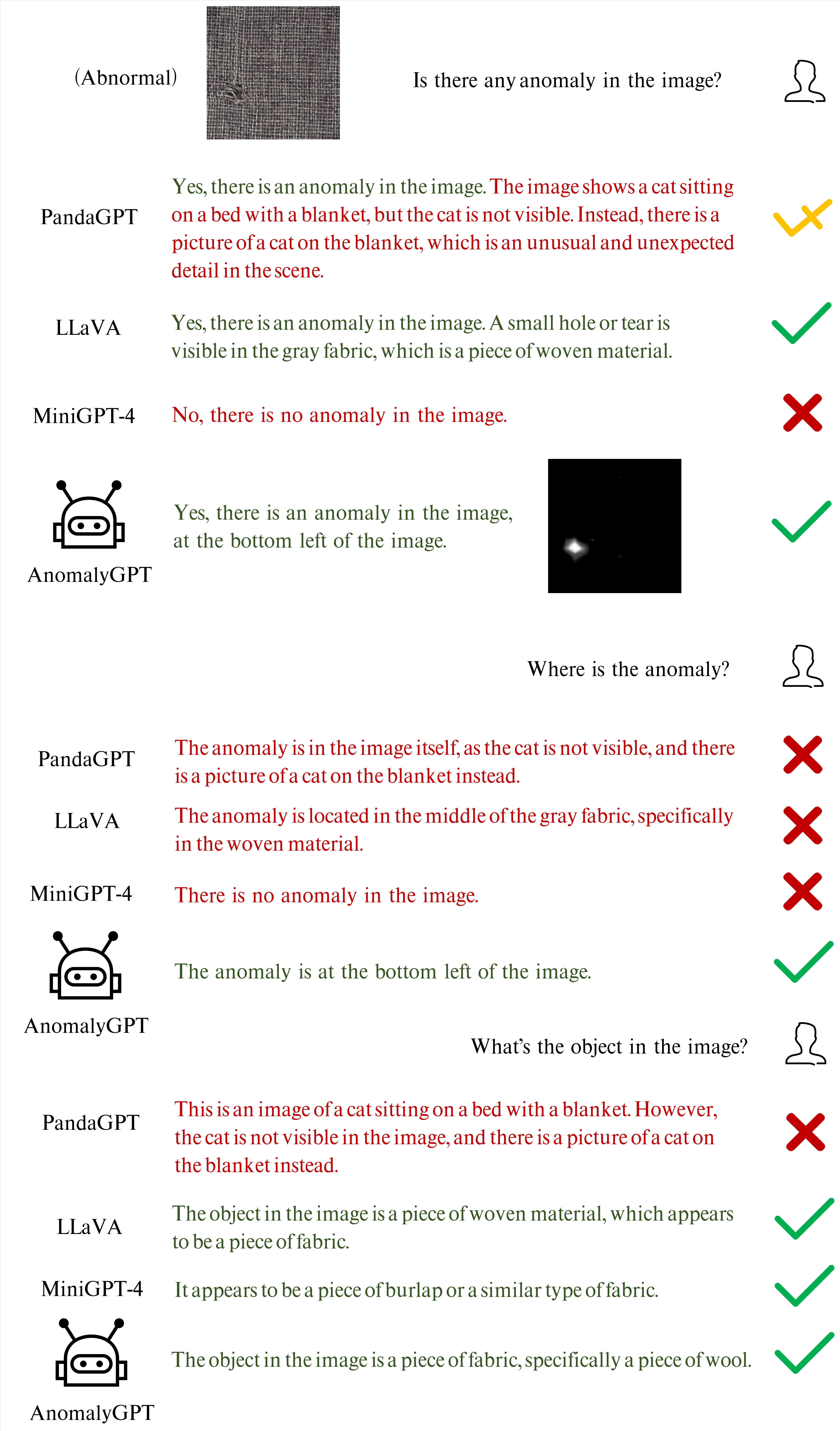} 
  \caption{Comparison between AnomalyGPT, PandaGPT, LLaVA and MiniGPT-4. The input image is a photo of a piece of farbic with hole. AnomalyGPT is capable of detecting anomaly, pinpointing its location, providing pixel-level localization results and answering questions about the image.}
  \label{fig:case_3}
\end{figure}

\begin{figure}[h]
  \centering
  \includegraphics[width=0.75\columnwidth]{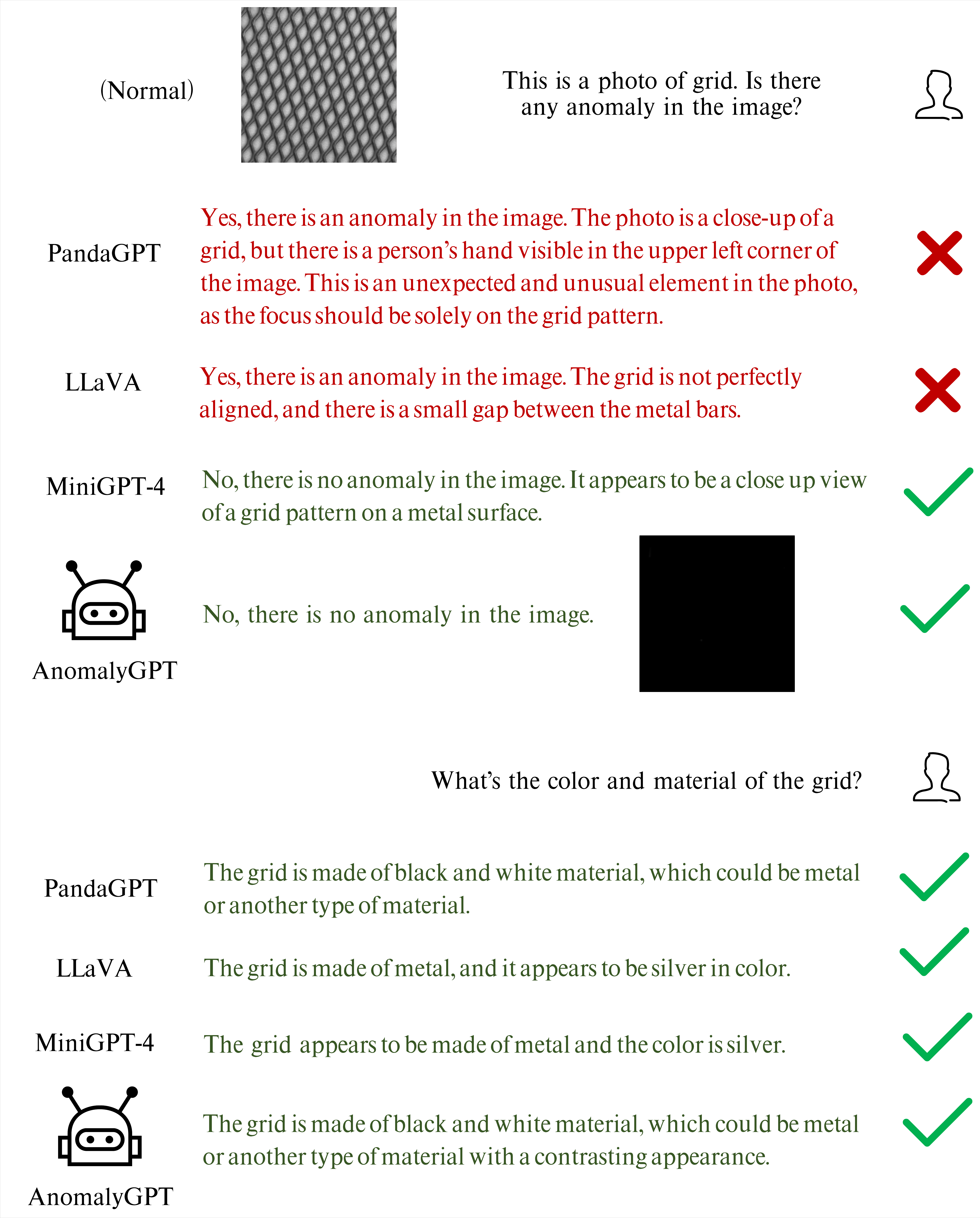} 
  \caption{Comparison between AnomalyGPT, PandaGPT, LLaVA and MiniGPT-4. The input image is a photo of normal metal grid. AnomalyGPT is capable of detecting anomaly, pinpointing its location, providing pixel-level localization results and answering questions about the image.}
  \label{fig:case_4}
\end{figure}

\begin{figure}[h]
  \centering
  \includegraphics[width=0.75\columnwidth]{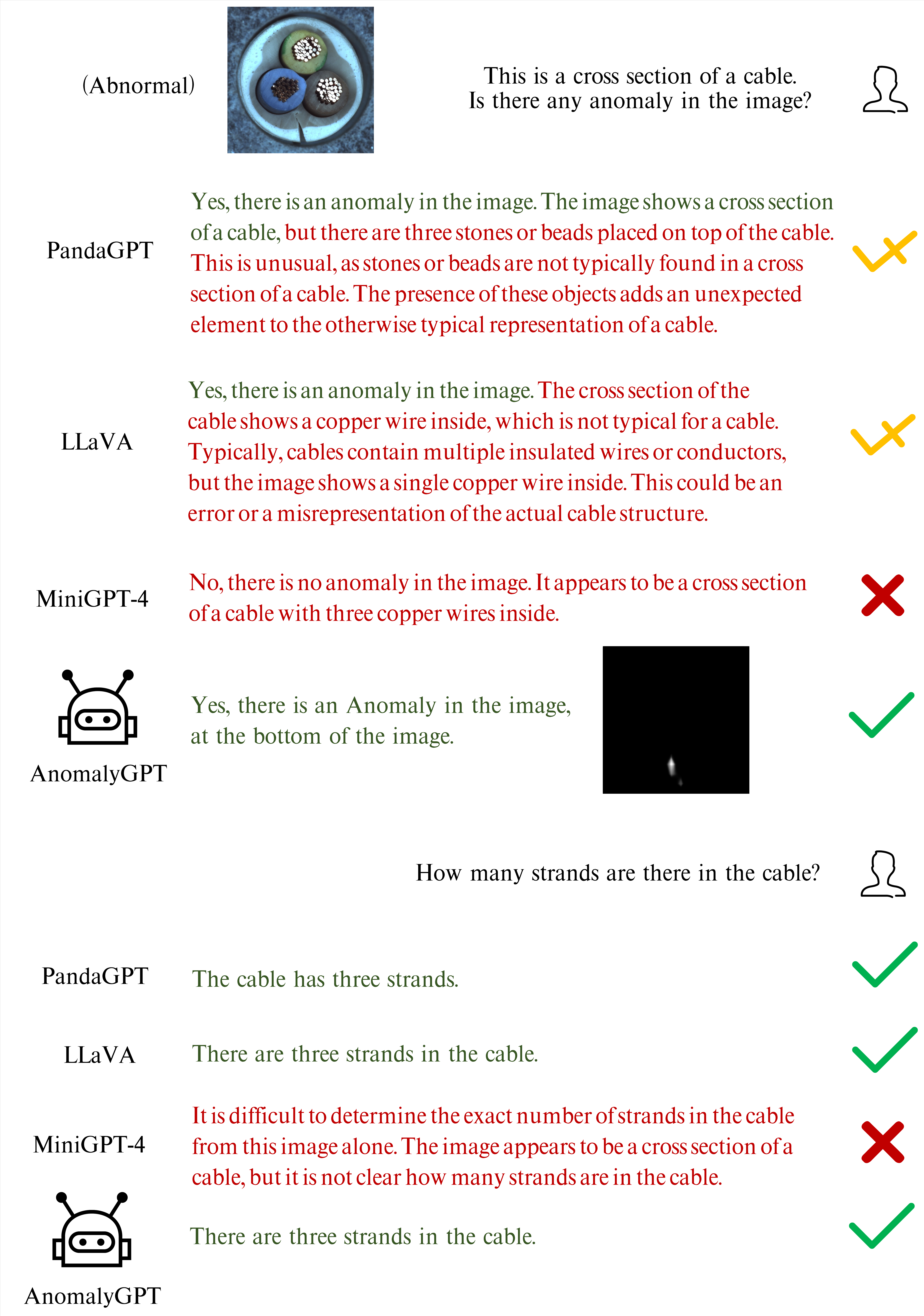} 
  \caption{Comparison between AnomalyGPT, PandaGPT, LLaVA and MiniGPT-4. The input image is a photo of a cable with defect. AnomalyGPT is capable of detecting anomaly, pinpointing its location, providing pixel-level localization results and answering questions about the image.}
  \label{fig:case_5}
\end{figure}

We compare our approach with several existing LVLMs, specifically selecting PandaGPT~\cite{su2023pandagpt}, MiniGPT-4~\cite{zhu2023minigpt}, and LLaVA~\cite{liu2023visual} for comparative analysis. We conduct experiments across various categories of both normal and anomalous samples. The results are presented in Figure~\ref{fig:case_c1}, Figure~\ref{fig:case_c2}, Figure~\ref{fig:case_6}, Figure~\ref{fig:case_3}, Figure~\ref{fig:case_4} and Figure~\ref{fig:case_5}. It can be observed that only our method exhibits proficiency in both accurately answering questions related to anomaly detection and those about image content. In contrast, the other models demonstrate suboptimal performance in discerning the presence of anomalies and pinpointing their precise locations. Notably, PandaGPT and LLaVA show a marked tendency to misclassify all samples as anomalous. Conversely, MiniGPT-4 tends to err on the side of caution, predominantly labeling samples as normal.

\end{document}

%% file: preamble.tex
\usepackage[dvipsnames]{xcolor}


%% file: main.bbl
\begin{thebibliography}{37}
\providecommand{\natexlab}[1]{#1}
\providecommand{\url}[1]{\texttt{#1}}
\expandafter\ifx\csname urlstyle\endcsname\relax
  \providecommand{\doi}[1]{doi: #1}\else
  \providecommand{\doi}{doi: \begingroup \urlstyle{rm}\Url}\fi

\bibitem[Bergmann et~al.(2019)Bergmann, Fauser, Sattlegger, and
  Steger]{bergmann2019mvtec}
Paul Bergmann, Michael Fauser, David Sattlegger, and Carsten Steger.
\newblock Mvtec ad--a comprehensive real-world dataset for unsupervised anomaly
  detection.
\newblock In \emph{Proceedings of the IEEE/CVF conference on computer vision
  and pattern recognition}, pages 9592--9600, 2019.

\bibitem[Chen et~al.(2023)Chen, Han, and Zhang]{chen2023zero}
Xuhai Chen, Yue Han, and Jiangning Zhang.
\newblock A zero-/few-shot anomaly classification and segmentation method for
  cvpr 2023 vand workshop challenge tracks 1\&2: 1st place on zero-shot ad and
  4th place on few-shot ad.
\newblock \emph{arXiv preprint arXiv:2305.17382}, 2023.

\bibitem[Chiang et~al.(2023)Chiang, Li, Lin, Sheng, Wu, Zhang, Zheng, Zhuang,
  Zhuang, Gonzalez, et~al.]{chiang2023vicuna}
Wei-Lin Chiang, Zhuohan Li, Zi Lin, Ying Sheng, Zhanghao Wu, Hao Zhang, Lianmin
  Zheng, Siyuan Zhuang, Yonghao Zhuang, Joseph~E Gonzalez, et~al.
\newblock Vicuna: An open-source chatbot impressing gpt-4 with 90\%* chatgpt
  quality.
\newblock \emph{See https://vicuna. lmsys. org (accessed 14 April 2023)}, 2023.

\bibitem[Chung et~al.(2022)Chung, Hou, Longpre, Zoph, Tay, Fedus, Li, Wang,
  Dehghani, Brahma, et~al.]{chung2022scaling}
Hyung~Won Chung, Le Hou, Shayne Longpre, Barret Zoph, Yi Tay, William Fedus,
  Eric Li, Xuezhi Wang, Mostafa Dehghani, Siddhartha Brahma, et~al.
\newblock Scaling instruction-finetuned language models.
\newblock \emph{arXiv preprint arXiv:2210.11416}, 2022.

\bibitem[Cohen and Hoshen(2020)]{cohen2020sub}
Niv Cohen and Yedid Hoshen.
\newblock Sub-image anomaly detection with deep pyramid correspondences.
\newblock \emph{arXiv preprint arXiv:2005.02357}, 2020.

\bibitem[Defard et~al.(2021)Defard, Setkov, Loesch, and
  Audigier]{defard2021padim}
Thomas Defard, Aleksandr Setkov, Angelique Loesch, and Romaric Audigier.
\newblock Padim: a patch distribution modeling framework for anomaly detection
  and localization.
\newblock In \emph{International Conference on Pattern Recognition}, pages
  475--489. Springer, 2021.

\bibitem[Dosovitskiy et~al.(2020)Dosovitskiy, Beyer, Kolesnikov, Weissenborn,
  Zhai, Unterthiner, Dehghani, Minderer, Heigold, Gelly,
  et~al.]{dosovitskiy2020image}
Alexey Dosovitskiy, Lucas Beyer, Alexander Kolesnikov, Dirk Weissenborn,
  Xiaohua Zhai, Thomas Unterthiner, Mostafa Dehghani, Matthias Minderer, Georg
  Heigold, Sylvain Gelly, et~al.
\newblock An image is worth 16x16 words: Transformers for image recognition at
  scale.
\newblock \emph{arXiv preprint arXiv:2010.11929}, 2020.

\bibitem[Girdhar et~al.(2023)Girdhar, El-Nouby, Liu, Singh, Alwala, Joulin, and
  Misra]{girdhar2023imagebind}
Rohit Girdhar, Alaaeldin El-Nouby, Zhuang Liu, Mannat Singh, Kalyan~Vasudev
  Alwala, Armand Joulin, and Ishan Misra.
\newblock Imagebind: One embedding space to bind them all.
\newblock In \emph{Proceedings of the IEEE/CVF Conference on Computer Vision
  and Pattern Recognition}, pages 15180--15190, 2023.

\bibitem[Gudovskiy et~al.(2022)Gudovskiy, Ishizaka, and
  Kozuka]{gudovskiy2022cflow}
Denis Gudovskiy, Shun Ishizaka, and Kazuki Kozuka.
\newblock Cflow-ad: Real-time unsupervised anomaly detection with localization
  via conditional normalizing flows.
\newblock In \emph{Proceedings of the IEEE/CVF Winter Conference on
  Applications of Computer Vision}, pages 98--107, 2022.

\bibitem[Huang et~al.(2022)Huang, Guan, Jiang, Zhang, Spratling, and
  Wang]{huang2022registration}
Chaoqin Huang, Haoyan Guan, Aofan Jiang, Ya Zhang, Michael Spratling, and
  Yan-Feng Wang.
\newblock Registration based few-shot anomaly detection.
\newblock In \emph{European Conference on Computer Vision}, pages 303--319.
  Springer, 2022.

\bibitem[Jeong et~al.(2023)Jeong, Zou, Kim, Zhang, Ravichandran, and
  Dabeer]{jeong2023winclip}
Jongheon Jeong, Yang Zou, Taewan Kim, Dongqing Zhang, Avinash Ravichandran, and
  Onkar Dabeer.
\newblock Winclip: Zero-/few-shot anomaly classification and segmentation.
\newblock In \emph{Proceedings of the IEEE/CVF Conference on Computer Vision
  and Pattern Recognition}, pages 19606--19616, 2023.

\bibitem[Lee et~al.(2022)Lee, Lee, and Song]{lee2022cfa}
Sungwook Lee, Seunghyun Lee, and Byung~Cheol Song.
\newblock Cfa: Coupled-hypersphere-based feature adaptation for target-oriented
  anomaly localization.
\newblock \emph{IEEE Access}, 10:\penalty0 78446--78454, 2022.

\bibitem[Lei et~al.(2023)Lei, Hu, Wang, and Liu]{lei2023pyramidflow}
Jiarui Lei, Xiaobo Hu, Yue Wang, and Dong Liu.
\newblock Pyramidflow: High-resolution defect contrastive localization using
  pyramid normalizing flow.
\newblock In \emph{Proceedings of the IEEE/CVF Conference on Computer Vision
  and Pattern Recognition}, pages 14143--14152, 2023.

\bibitem[Li et~al.(2021)Li, Sohn, Yoon, and Pfister]{li2021cutpaste}
Chun-Liang Li, Kihyuk Sohn, Jinsung Yoon, and Tomas Pfister.
\newblock Cutpaste: Self-supervised learning for anomaly detection and
  localization.
\newblock In \emph{Proceedings of the IEEE/CVF conference on computer vision
  and pattern recognition}, pages 9664--9674, 2021.

\bibitem[Li et~al.(2023)Li, Li, Savarese, and Hoi]{li2023blip}
Junnan Li, Dongxu Li, Silvio Savarese, and Steven Hoi.
\newblock Blip-2: Bootstrapping language-image pre-training with frozen image
  encoders and large language models.
\newblock \emph{arXiv preprint arXiv:2301.12597}, 2023.

\bibitem[Lin et~al.(2017)Lin, Goyal, Girshick, He, and
  Doll{\'a}r]{lin2017focal}
Tsung-Yi Lin, Priya Goyal, Ross Girshick, Kaiming He, and Piotr Doll{\'a}r.
\newblock Focal loss for dense object detection.
\newblock In \emph{Proceedings of the IEEE international conference on computer
  vision}, pages 2980--2988, 2017.

\bibitem[Liu et~al.(2023)Liu, Li, Wu, and Lee]{liu2023visual}
Haotian Liu, Chunyuan Li, Qingyang Wu, and Yong~Jae Lee.
\newblock Visual instruction tuning.
\newblock \emph{arXiv preprint arXiv:2304.08485}, 2023.

\bibitem[Milletari et~al.(2016)Milletari, Navab, and Ahmadi]{milletari2016v}
Fausto Milletari, Nassir Navab, and Seyed-Ahmad Ahmadi.
\newblock V-net: Fully convolutional neural networks for volumetric medical
  image segmentation.
\newblock In \emph{2016 fourth international conference on 3D vision (3DV)},
  pages 565--571. Ieee, 2016.

\bibitem[Ouyang et~al.(2022)Ouyang, Wu, Jiang, Almeida, Wainwright, Mishkin,
  Zhang, Agarwal, Slama, Ray, et~al.]{ouyang2022training}
Long Ouyang, Jeffrey Wu, Xu Jiang, Diogo Almeida, Carroll Wainwright, Pamela
  Mishkin, Chong Zhang, Sandhini Agarwal, Katarina Slama, Alex Ray, et~al.
\newblock Training language models to follow instructions with human feedback.
\newblock \emph{Advances in Neural Information Processing Systems},
  35:\penalty0 27730--27744, 2022.

\bibitem[P{\'e}rez et~al.(2003)P{\'e}rez, Gangnet, and Blake]{perez2003poisson}
Patrick P{\'e}rez, Michel Gangnet, and Andrew Blake.
\newblock Poisson image editing.
\newblock In \emph{ACM SIGGRAPH 2003 Papers}, pages 313--318. 2003.

\bibitem[Pirnay and Chai(2022)]{pirnay2022inpainting}
Jonathan Pirnay and Keng Chai.
\newblock Inpainting transformer for anomaly detection.
\newblock In \emph{International Conference on Image Analysis and Processing},
  pages 394--406. Springer, 2022.

\bibitem[Radford et~al.(2021)Radford, Kim, Hallacy, Ramesh, Goh, Agarwal,
  Sastry, Askell, Mishkin, Clark, et~al.]{radford2021learning}
Alec Radford, Jong~Wook Kim, Chris Hallacy, Aditya Ramesh, Gabriel Goh,
  Sandhini Agarwal, Girish Sastry, Amanda Askell, Pamela Mishkin, Jack Clark,
  et~al.
\newblock Learning transferable visual models from natural language
  supervision.
\newblock In \emph{International conference on machine learning}, pages
  8748--8763. PMLR, 2021.

\bibitem[Roth et~al.(2022)Roth, Pemula, Zepeda, Sch{\"o}lkopf, Brox, and
  Gehler]{roth2022towards}
Karsten Roth, Latha Pemula, Joaquin Zepeda, Bernhard Sch{\"o}lkopf, Thomas
  Brox, and Peter Gehler.
\newblock Towards total recall in industrial anomaly detection.
\newblock In \emph{Proceedings of the IEEE/CVF Conference on Computer Vision
  and Pattern Recognition}, pages 14318--14328, 2022.

\bibitem[Schl{\"u}ter et~al.(2022)Schl{\"u}ter, Tan, Hou, and
  Kainz]{schluter2022natural}
Hannah~M Schl{\"u}ter, Jeremy Tan, Benjamin Hou, and Bernhard Kainz.
\newblock Natural synthetic anomalies for self-supervised anomaly detection and
  localization.
\newblock In \emph{European Conference on Computer Vision}, pages 474--489.
  Springer, 2022.

\bibitem[Su et~al.(2023)Su, Lan, Li, Xu, Wang, and Cai]{su2023pandagpt}
Yixuan Su, Tian Lan, Huayang Li, Jialu Xu, Yan Wang, and Deng Cai.
\newblock Pandagpt: One model to instruction-follow them all.
\newblock \emph{arXiv preprint arXiv:2305.16355}, 2023.

\bibitem[Touvron et~al.(2023)Touvron, Lavril, Izacard, Martinet, Lachaux,
  Lacroix, Rozi{\`e}re, Goyal, Hambro, Azhar, et~al.]{touvron2023llama}
Hugo Touvron, Thibaut Lavril, Gautier Izacard, Xavier Martinet, Marie-Anne
  Lachaux, Timoth{\'e}e Lacroix, Baptiste Rozi{\`e}re, Naman Goyal, Eric
  Hambro, Faisal Azhar, et~al.
\newblock Llama: Open and efficient foundation language models.
\newblock \emph{arXiv preprint arXiv:2302.13971}, 2023.

\bibitem[Wang et~al.(2023)Wang, Chen, Chen, Wu, Zhu, Zeng, Luo, Lu, Zhou, Qiao,
  et~al.]{wang2023visionllm}
Wenhai Wang, Zhe Chen, Xiaokang Chen, Jiannan Wu, Xizhou Zhu, Gang Zeng, Ping
  Luo, Tong Lu, Jie Zhou, Yu Qiao, et~al.
\newblock Visionllm: Large language model is also an open-ended decoder for
  vision-centric tasks.
\newblock \emph{arXiv preprint arXiv:2305.11175}, 2023.

\bibitem[Wyatt et~al.(2022)Wyatt, Leach, Schmon, and
  Willcocks]{wyatt2022anoddpm}
Julian Wyatt, Adam Leach, Sebastian~M Schmon, and Chris~G Willcocks.
\newblock Anoddpm: Anomaly detection with denoising diffusion probabilistic
  models using simplex noise.
\newblock In \emph{Proceedings of the IEEE/CVF Conference on Computer Vision
  and Pattern Recognition}, pages 650--656, 2022.

\bibitem[Xie et~al.(2023)Xie, Wang, Liu, Zheng, and Jin]{xie2023pushing}
Guoyang Xie, Jingbao Wang, Jiaqi Liu, Feng Zheng, and Yaochu Jin.
\newblock Pushing the limits of fewshot anomaly detection in industry vision:
  Graphcore.
\newblock \emph{arXiv preprint arXiv:2301.12082}, 2023.

\bibitem[Yan et~al.(2021)Yan, Zhang, Xu, Hu, and Heng]{yan2021learning}
Xudong Yan, Huaidong Zhang, Xuemiao Xu, Xiaowei Hu, and Pheng-Ann Heng.
\newblock Learning semantic context from normal samples for unsupervised
  anomaly detection.
\newblock In \emph{Proceedings of the AAAI conference on artificial
  intelligence}, pages 3110--3118, 2021.

\bibitem[Yi and Yoon(2020)]{yi2020patch}
Jihun Yi and Sungroh Yoon.
\newblock Patch svdd: Patch-level svdd for anomaly detection and segmentation.
\newblock In \emph{Proceedings of the Asian conference on computer vision},
  2020.

\bibitem[You et~al.(2022)You, Cui, Shen, Yang, Lu, Zheng, and
  Le]{you2022unified}
Zhiyuan You, Lei Cui, Yujun Shen, Kai Yang, Xin Lu, Yu Zheng, and Xinyi Le.
\newblock A unified model for multi-class anomaly detection.
\newblock \emph{Advances in Neural Information Processing Systems},
  35:\penalty0 4571--4584, 2022.

\bibitem[Zavrtanik et~al.(2021)Zavrtanik, Kristan, and
  Sko{\v{c}}aj]{zavrtanik2021reconstruction}
Vitjan Zavrtanik, Matej Kristan, and Danijel Sko{\v{c}}aj.
\newblock Reconstruction by inpainting for visual anomaly detection.
\newblock \emph{Pattern Recognition}, 112:\penalty0 107706, 2021.

\bibitem[Zhao(2022)]{zhao2022just}
Ying Zhao.
\newblock Just noticeable learning for unsupervised anomaly localization and
  detection.
\newblock In \emph{2022 IEEE International Conference on Multimedia and Expo
  (ICME)}, pages 01--06. IEEE, 2022.

\bibitem[Zhao(2023)]{zhao2023omnial}
Ying Zhao.
\newblock Omnial: A unified cnn framework for unsupervised anomaly
  localization.
\newblock In \emph{Proceedings of the IEEE/CVF Conference on Computer Vision
  and Pattern Recognition}, pages 3924--3933, 2023.

\bibitem[Zhu et~al.(2023)Zhu, Chen, Shen, Li, and Elhoseiny]{zhu2023minigpt}
Deyao Zhu, Jun Chen, Xiaoqian Shen, Xiang Li, and Mohamed Elhoseiny.
\newblock Minigpt-4: Enhancing vision-language understanding with advanced
  large language models.
\newblock \emph{arXiv preprint arXiv:2304.10592}, 2023.

\bibitem[Zou et~al.(2022)Zou, Jeong, Pemula, Zhang, and Dabeer]{zou2022spot}
Yang Zou, Jongheon Jeong, Latha Pemula, Dongqing Zhang, and Onkar Dabeer.
\newblock Spot-the-difference self-supervised pre-training for anomaly
  detection and segmentation.
\newblock In \emph{European Conference on Computer Vision}, pages 392--408.
  Springer, 2022.

\end{thebibliography}
